\documentclass[a4paper,fleqn]{cas-dc}

\usepackage[numbers,sort&compress]{natbib}
\usepackage{subfigure}
\usepackage{multirow}
\usepackage{graphicx}
\usepackage{placeins}
\usepackage{stfloats}
\usepackage{url}
\usepackage{cuted}\stripsep -3pt plus 3pt minus 2pt
\usepackage{caption}
\usepackage{booktabs}
\usepackage{algorithm}
\usepackage{algorithmicx}
\usepackage{algpseudocode}



\def\tsc#1{\csdef{#1}{\textsc{\lowercase{#1}}\xspace}}
\tsc{WGM}
\tsc{QE}
\tsc{EP}
\tsc{PMS}
\tsc{BEC}
\tsc{DE}

\begin{document}
\let\WriteBookmarks\relax
\def\floatpagepagefraction{l}
\def\textpagefraction{.001}
\shorttitle{Self-supervised asymmetric deep hashing with margin-scalable constraint}
\shortauthors{Z. Yu et~al.}

\title [mode = title]{Self-supervised asymmetric deep hashing with margin-scalable constraint}                      




\author[1]{Zhengyang Yu}
\fnmark[1]


\address[1]{College of Computer and Information Science, Southwest University, Chongqing, China}

\author[1]{Song Wu}
\cormark[1]
\cortext[cor1]{Corresponding author}
\ead{songwuswu@swu.edu.cn}

\author[1]{Zhihao Dou}
\fnmark[2]

\author[2]{Erwin M. Bakker}
\address[2]{LIACS Media Lab, Leiden University, Leiden, Netherlands}









\begin{abstract}
  Due to its effectivity and efficiency, deep hashing approaches are widely used for large-scale visual search. However, it is still challenging to produce compact and discriminative hash codes for images associated with multiple semantics for two main reasons, 1) similarity constraints designed in most of the existing methods are based upon an oversimplified similarity assignment(i.e., 0 for instance pairs sharing no label, 1 for instance pairs sharing at least 1 label), 2) the exploration in multi-semantic relevance are insufficient or even neglected in many of the existing methods. These problems significantly limit the discrimination of generated hash codes. In this paper, we propose a novel self-supervised asymmetric deep hashing method with a margin-scalable constraint(SADH) approach to cope with these problems. SADH implements a self-supervised network to sufficiently preserve semantic information in a semantic feature dictionary and a semantic code dictionary for the semantics of the given dataset, which efficiently and precisely guides a feature learning network to preserve multi-label semantic information using an asymmetric learning strategy. By further exploiting semantic dictionaries, a new margin-scalable constraint is employed for both precise similarity searching and robust hash code generation. Extensive empirical research on four popular benchmarks validates the proposed method and shows it outperforms several state-of-the-art approaches. The source codes URL of our SADH is: \url{http://github.com/SWU-CS-MediaLab/SADH}.
\end{abstract}



\begin{keywords}
  Deep supervised hashing\sep Asymmetric learning\sep Self-supervised learning
\end{keywords}

\maketitle

\section{Introduction}
The amount of image and video data in social networks and search engines are growing at an alarming rate. In order to effectively search large-scale high dimensional image data,  Approximate Nearest Neighbor (ANN) search has been extensively studied by researchers\cite{1,2}. Semantic hashing, first proposed in the pioneer work\cite{100} is widely used in the field of large-scale image retrieval. It maps high-dimensional content features of pictures into Hamming space (binary spa- ce) to generate a low-dimensional hash sequence\cite{1,2}, which reflects the semantic similarity by distance between hash cod- es in the Hamming space.
Hash algorithms can be broadly divided into data-dependent methods and data-independent methods\cite{31} schemes. The most basic but representative data independent method is Locality Sensitive Hashing LSH\cite{1}, which generates embedding through random projections. Ho- wever, these methods all require long binary code to achieve accuracy, which is not adapt to the processing of large-scale visual data. Recent research priorities have shifted to data-dependent approaches that can generate compact binary codes by learning large amount of data and information. This type of method embeds high-dimensional data into the Hamming space and performs bitwise operations to find similar objects. Recent data-dependent works such as \cite{2,5,6,7,8,59,60} have shown better retrieval accuracy under smaller hash code len- gth.

Although the above data-dependent hashing methods ha-ve certainly succeeded to some extent, they all use hand-crafted features, thereby limiting the retrieval accuracy of learning binary code. Recently, the deep-learning-based has-hing methods have shown superior performance by combining the powerful feature extraction of deep learning\cite{9,25,11,12,13,14}.
Admitting significant progress achieved in large-scale image retrieval with deep hashing methods, there still remain crucial bottlenecks that limit the hashing retrieval accuracy for datasets like NUS-WIDE\cite{15}, MS-COCO\cite{68}, MIRFlickr-25K\cite{16}, where each image is annotated with multiple semantics. Firstly, to the best of our knowledge, most of the existing supervised hashing methods use semantic-level labels to examine the similarity between instance pairs following a common experimental protocol. That is, the similarity score will be assigned as ‘1’ if the item pair shares at least one semantic label and ‘0’ if none of the semantic labels are shared. Based upon this coarsely defined similarity metric, in many of the existing methods\cite{27,9,47}, the exact degree of similarity(i.e., how many exact semantics are shared) cannot be quantified, therefore they fail to search for similarity information at a fine-grained level. Additionally, by further utilizing semantic labels, exploring semantic relevance to facilitate the similarity searching process can bring numerous merits for hashing function learning, e.g., the inter-class instance pairs can be better separated which can provide better efficiency and robustness in the training process\cite{67}; the shared image representations can be learned which is beneficial for hashing function learning\cite{42}. Many existing deep hashing methods ignore to leverage such valuable semantic information\cite{9,25,11,13,14}, leading to inferior retrieval performance. A few of the existing methods\cite{28,65,66,42} solve this problem by adding an auxiliary classifier to enhance the preservation of global semantic information. However, the complex semantic correlations under mentioned multi-label scenarios are still insufficiently discovered and cannot be effectively embedded into hash codes.

To tackle the mentioned flaws, we proposed a novel self-supervised asymmetric deep hashing with margin-scalable constraint(SADH) approach to improve the accuracy and efficiency of multi-label image retrieval. Holding the motivation of thoroughly discover semantic relevance, as shown in Fig.1, in our work, in spite of using an auxiliary classifier following methods like\cite{28,65,66,42}, semantic relevance from multi-label annotations are thoroughly excavated through a self-supervised Semantic-Network. While a convolutional neural network namely Image-Network, projects original image inputs into semantic features and hash codes. Inspired by methods like\cite{88,89,90,32}, we propose a novel asymmetric guidance mechanism to efficiently and effectively transfer semantic information from Semantic-Network to Image-Network, firstly we refine the abstract semantic features and binary codes of the entire training set labels generated by Semantic-Network into two semantic dictionaries by removing the duplications, by which the global knowledge stored in semantic dictionaries can seamlessly supervise the feature learning and hashing generation of Image-Network for each sampled mini-batch of input images with asymmetric association. Additionally, we are also motivated to search pairwise similarity at a fine-grained level. To this end, a well-defined margin-scalable pairwise constraint is proposed. Unlike conventional similarity constraint used in many existing metho- ds\cite{27,9,47} with which all the similarity instance pairs are penalized with the same strength, by looking up the semantic dictionaries, our margin-scalable constraint can dynamically penalize instance pairs with respect to their corresponding semantic similarity in fine-grained level(i.e., for a given similarity score of one instance pair, the more identical semantics they share, the larger penalty would be given on them), with which our SADH is empowered to search for discriminative visual feature representations and corresponding combat hashing representations. The main contributions of this paper are as follows:

\begin{itemize}
    \item[1)] We propose a novel end-to-end deep hashing framework which consists of Image-Network and Semantic-Network. With a novel asymmetric guidance mechanism, rich semantic information preserved by Semantic-Network can be seamlessly transferred to Image-Netw-ork, which can ensure that the global semantic relevance can be sufficiently discovered and utilized from multi-label annotations of the entire training set. 
     
    \item[2)] We devise a novel margin-scalable pairwise constraint based upon the semantic dictionaries, which can effectively search for precise pairwise similarity information in a semantically fine-grained level to facilitate the discrimination of generated hash codes.
     
    \item[3)] Without losing generality, we comprehensively evaluate our proposed method on CIFAR-10, NUS-WIDE, MS-COCO, and MIRFlickr-25K to cope with image retrieval task, the effectiveness of proposed modules in our method is endorsed by exhaustive ablation studies. Additionally, we show how to seamlessly extend our SADH algorithm from single-modal scenario to multi-modal scenario. Extensive experiments demonstrate the superiority of our SADH in both image retrieval and cross-modal retrieval, as compared with several state-of-the-art hashing methods. 
\end{itemize}

\section{Related work}
In this section, we discuss works that are inspiring for our SADH or relevant to four popular research topics in learning to hash.

\subsection{Unsupervised hashing methods}
The unsupervised hashing methods endeavors to learn a set of hashing functions without any supervised information, they preserve the geometric structure (e.g., the similarity between neighboring samples) of the original data space, by which instance pairs that are close in the original data space are projected into similar hash codes, while the separated pairs in the original data space are projected into dissimilar hash codes. Locality sensitive hashing is the pioneer work of unsupervised hashing, which is first proposed in \cite{69, 70}, the basic idea of LSH is to learn a family of hashing functions that assigns similar item pairs with a higher probability of being mapped into the same hash code than dissimilar ones. Following \cite{69, 70}, many variants of LSH has been proposed, e.g., \cite{71, 72, 73} extends LSH from the traditional vector-to-vector nearest neighbor search to subspace-to-subspace nearest neighbor search with angular distance as subspace similarity metric. Although LSH can effectively balance computational cost and retrieval accuracy, but it has no exploration on the specific data distributions and often reveals inferior performance. In this paper, we focus on the data-dependent(learning to hash methods). The representative unsupervised learning to hash method includes ITQ\cite{4} which is the first method that learns relaxed hash codes with principal component analysis and iteratively minimize the quantization loss. SH\cite{8} proves the problem of finding good binary code for a given dataset is equivalent to the NP-hard graph partitioning problem, then the spectral relaxation scheme of the original problem is solved by identify the eigenvector solution. LSMH\cite{74} utilizes matrix decomposition to refine the original feature space into a latent feature space which makes both the latent features and binary codes more discriminative, this simultaneous feature learning and hashing learning scheme is followed by many latter methods.  
\subsection{Supervised hashing methods}
The supervised hashing methods can use the available supervised information such as labels or semantic affinities to guide feature extraction and hash code generation, which can achieve more robust retrieval performance than unsupervised methods. Supervised hashing with kernel (KSH) \cite{6} and supervised discrete hashing (SDH) \cite{19} generate binary hash codes by minimizing the Hamming distance through similar data point pairs. Distortion Minimization Hashing (DMS)\cite{59}, Minimum Loss Hashing (MLH)\cite{21}. Binary Reconstruction Embedding(BRE)\cite{59} learns hashing function by minimizing the reconstruction loss to similarities in the original feature space and Hamming space. In \cite{85,86}, Support Vector Machine(SVM) is used to learn a set of hyperplanes as a hash function family, by which the margin between the selected support vectors belonging to similar and dissimilar pairs are maximized to generate discriminative binary codes.  
Although the above hashing methods have certainly succeeded to some extent, they all use hand-crafted features that do not fully capture the semantic information and cannot search for similarity information in latent feature space and Hamming space simultaneously, thereby causing suboptimal problem.
Recently, the deep learning-based hashing methods have shown superior performance by exploiting the powerful feature extraction of deep learning\cite{21,23,24,53,54,55,56,57,58,add55,add56,add57}. In particular, Convolutional Neural Network Hash (CNNH) \cite{42} is a two-stage hashing method, where the pairwise similarity matrix is decomposed to approximate the optimal hash code representations which can directly guide hash function learning. However, in the two-stage framework of CNNH, the generation of latent features are not participated in the generation of approximate hash codes, so it fails to perform simultaneous feature extraction and hash code learning which limit the discrimination of hash codes. To solve this limitation, Yan et al \cite{74} improved \cite{42} by equally dividing the latent features into pieces then projecting the pieces of features into the bit-wise representations of hash codes under a one-stage framework. Similarly DSPH\cite{9} performs joint hash code learning and feature learning under a one-stage framework. DDSH\cite{27} adopt an alternative training strategy to optimize the continuous features and binary codes individually.

Although these methods have obtained satisfactory retrieval performance, they are still suboptimal for multi-label datasets, as they fail to sufficiently discover semantic relevance from multi-label annotations, additionally they only utilize the earlier mentioned coarsely defined similarity supervision(either 0 or 1), which fails to construct more precise pairwise correlations between pairs of hash codes and deep features, significantly downgrading retrieval accuracy. As stated by\cite{75}, multi-label images are widely involved in many large-scaled image retrieval systems, so it is valuable to improve the retrieval performance under this scenario. Many recent works are proposed which aim to fully exploit semantic labels in hash function learning. One natural and popular strategy used in a number of recent methods like\cite{28, 76, 77, 78, 79, 80, 81} is to add an auxiliary classifier in the hashing network to learn the hashing task and classification task simultaneously, which can provide more robust hash function learning by preserving semantic-specific features. A novel and effective methods DSEH\cite{29} utilizes a self-supervised semantic network to capture rich semantic information from semantic labels to guide the feature learning network which learns hash function for images. In comparison with auxiliary classifiers used in \cite{28, 76, 77, 78, 79, 80, 81}, the Semantic-Network used in DESH\cite{29} can capture more complex semantic correlations and can directly supervise the hash code generation, which significantly improves the retrieval performance in multi-label scenarios, however DSEH uses a conventional negative log-likelihood objective function which still cannot search for similarity information in a fine-grained level. Several methods design weighted ranking loss to solve this problem, e.g., HashNet\cite{12} tackle the ill-posed gradient problem of learning discrete hash function by changing the widely used negative log-likelihood objective function\cite{29, 9} into a Weighted Maximum Likelihood(WML) estimation. Yan et al. propose an instance-aware hashing framework for multi-label image retrieval in \cite{75}, where a weighted triplet loss is included based upon multi-label annotations. Similarly, DSRH\cite{82} designs a Surrogate Loss, in which a dynamic weight factor namely Normalized Discounted Cumulative Gain (NDCG) score is calculated which is related to the instance pairs’ shared number of labels. However, since both \cite{75} and \cite{82} design their weighted ranking loss in triplet form, they only consider preserving correct ranking of instances, instead of directly optimizing the multi-level pairwise semantic similarity. IDHN\cite{83} calculate a soft semantic similarity score(i.e., the cosine similarity between label pairs) to replace the hard-assigned semantic similarity metric, which directly perform as the supervision of negative log-likelihood pairwise loss. Although the soft semantic similarity score used in IDHN and the weight factor used in \cite{75}, \cite{12} and \cite{82} can reflect multi-level semantic similarity between labels, but they cannot guarantee that the predefined similarity measurement such as NDCG and cosine similarity is the optimal choice for supervising similarity searching of hash codes. 

Unlike these methods, we design a new similarity constraint in a contrastive form\cite{84}, which contains a margin parameter which can reflect the strength of supervision given on instance pairs. Inspired by DSEH\cite{29}, we observe that, using a self-supervi- sed training scheme and taking semantic labels as inputs, Semantic-Network can generate highly discriminative hash codes and its retrieval performance is not sensitive to the selection of hyper-parameter. Taking advantage of these characteristics of Semantic-Nework, we consider the pairwise similarity preserved by Semantic-Net- work as the optimum of an ideal hash function, by calculating a scalable margin factor for each item pairs with respect to the corresponding semantic information stored by Semantic-Network, our new similarity constraint can dynamically and accurately penalize the item pairs with respect to multi-level semantic similarity to learn combat hash codes. Note that the margin used in our method is originated form \cite{84}, this is different from the hyperplane margin used in SVM-based methods like\cite{85, 86}, which is maximized between negative and positive support vectors. Additionally, a similar form of contrastive loss function can be also seen in MMHH\cite{87}, which also contains a margin value. However different from our SADH, which is mainly focus on multi-label image retrieval, MMHH is focused on alleviating the vulnerability to noisy data. In comparison with our scalable margin, the margin used in MMHH is fixed based on manual selection, which is viewed as Hamming radius to truncate the contrastive loss, preventing it from being excessively large for noisy data.

\subsection{Asymmetric hashing methods}
Most classical hashing methods build pairwise interaction in symmetric form, recently asymmetric hashing methods have shown the power of learning distinct hash functions and building asymmetric interactions in similarity search. Asymmetric LSH\cite{88} extends LSH to solve the approximate Maximum Inner Product Search (MIPS) problem by generalizing the MIPS problem to an ANN problem with asymmetric transformation. However, asymmetric LSH is data-independent and can hardly achieve satisfactory result. SSA- H\cite{62} directly solve the MIPS problem by approximating the full similarity matrix using asymmetric learning structure. \cite{90} theoretically interprets that there is an exponential gap between the minimal binary code length of symmetric and asymmetric hashing. NAMVH\cite{81} learns a real-valued non-linear embedding for novel query data and a multi-integer embedding for the entire database and correlate two distinct embedding asymmetrically. In the deep hashing framework ADSH\cite{32}, only query points are engaged in the stage of updating deep network parameters, while the hash codes for database are directly learned as a auxiliary variable, the hash codes generated by the query and database are correlated through asymmetric pairwise constraints, such that the dataset points can be efficiently utilized during the hash function learning procedure. In comparison with \cite{32} building asymmetric association between query and database, notably the cross-modal hashing framework AGAH\cite{34} is devoted to use the asymmetric learning strategy to fully preserve semantic relevance between multi-modal feature representations and their corresponding label information to eliminate modality gap,
It constructs asymmetric interaction between binary codes belonging to heterogeneous modalities and semantic labels. Different from AGAH, which separately learns hash function for each single semantics to build asymmetric interaction with modalities, our method leverage a self-supervised network to directly learn hash function for multi-label annotations, which can indicate more fine-grained similarity information. We preserve semantic information from labels of the entire training set, which in turn being refined in form of two semantic dictionaries. Comparing to DSEH\cite{29} which utilize an alternative training strategy and point-to-point symmetric supervision, with the asymmetric guidance of two dictionaries in our method, the global semantic relevance can be more powerfully and efficiently transferred to hash codes and latent feature generated by each sampled mini-batch of images. 

\subsection{Cross-modal hashing methods}
\label{sec_24}
Cross-modal hashing(CMH) has become an active research area since IMH\cite{91} extends the scenario of hashing from similarity search of traditional homogeneous data to heterogeneous data by exploring inter-and-intra consistency and projecting the multi-modality data to a common hamming space. Followed by which a number of CMH methods are proposed, representative unsupervised methods include LSSH\cite{92} which is the first CMH method that simultaneous do similarity search in latent feature space and Hamming space, CMFH\cite{93} uses collective matrix factorization to correlate different modalities and CVH\cite{94}which is the extension of SH for solving cross-view retrieval. Similar to single modal hashing, CMH can achieve more powerful performance with supervised information. SCM\cite{95} is the first attempt to integrate semantic labels into a CMH framework. SePH\cite{96} minimize the Kullback-Leibler(KL) divergence between the pairwise similarity of labels and hash codes. Recently, due to the powerful ability of deep learning in feature extraction, more and more efforts have been devoted to deep cross-modal hashing. Similar to DSPH\cite{9}, DCMH\cite{97} and PRDH\cite{98} performs simultaneous feature learning and hash learning under and end-to-end framework. The preservation of semantic relevance is also beneficial for bridging heterogeneous data. Multi-Task Consistency-Preserving Adversarial Hashing(CPAH) \cite{99} devise an adversarial module and classification module to align the feature distribution and semantic consistency between different modality data. SSAH\cite{62} utilize the self-supervised semantic network in a way that is similar to DSEH, to learn a common semantic space for different modalities. In this paper, although we mainly focus on the single-modal scenario, the core components of our SADH algorithm can be seamlessly integrated in a cross-modal hashing framework. The extension of our method from single-modal to multi-modal scenarios is discussed, and we demonstrate that our SADH can achieve state-of-the-art experimental performance in both scenarios.

\begin{figure*}
	\centering
  \includegraphics[width=\linewidth]{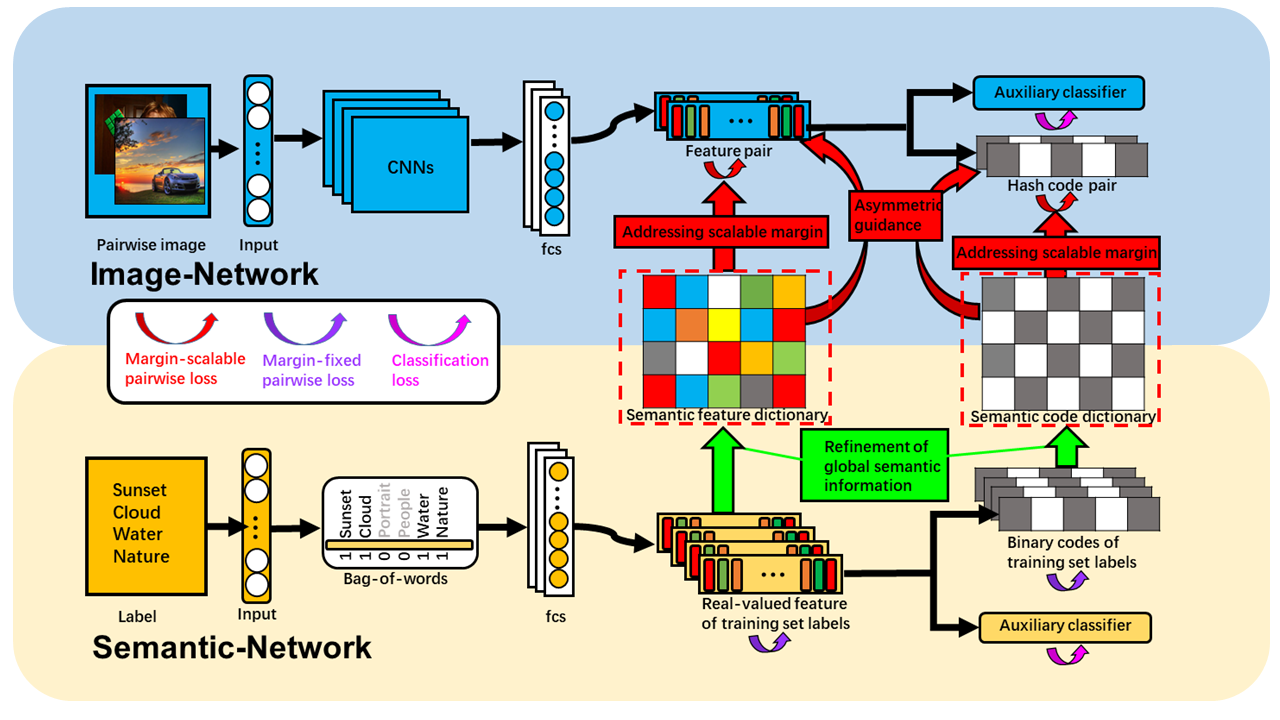}
  \caption{The overall framework of our proposed SADH, Image-Network plotted in blue background is comprised of CNN layers for deep image representations, while Semantic-Network plotted in yellow background is a self-supervised MLP network which abstracts semantic features from one-hot annotations as inputs. Both networks embeds deep features into a semantic space through a semantic layer, and independently obtain classification outputs and binary codes using multi-task learning framework. Semantic-Network is first trained until convergence, then global semantic information of the entire training set labels is refined by Semantic-Network into two semantic dictionaries, such refined semantic information is transferred to Image-Network by asymmetric guidance on both feature learning and hash code generation. The semantic dictionaries are further utilized to dynamically assign each instance pairs of Image-Network with a scalable margin in the pairwise constraint.}
	\label{fig_1}
\end{figure*}

\section{The proposed method}
We elaborate our proposed SADH in details. Firstly, the problem formulation for hash function learning is presented. Afterwards, each module as well as the optimization strategy in the Semantic-Network and Image-Network are explicitly described. As can be seen in the overall framework Fig. \ref{fig_1}, SADH consists of two networks, where Semantic-Network is a pure MLP network for semantic preservation with labels in form of bag-of-words as inputs. Image-Network utilizes convolutional neural network to extract high-dimensional visual feature from images, which in turn being projected into binary hash codes, with both deep features (generated by semantic layer) and hash codes (generated by hash layer) under asymmetric guidance of Semantic-Network as shown in Fig. \ref{fig_1}.

\subsection{Problem definition}
First the notations used in the rest of the paper are introduced. Following methods like\cite{29, 28, 19, 82}, we consider the common image retrieval scenario where images are annotated by semantic labels, let $O=\left\{o_{l}\right\}_{i=1}^{m}$ denote a dataset with m instances, and $o_{l}=\left(v_{l}, l_{l}\right)$ where $v_{l} \in \mathbb{R}^{1 \times d_{v}}$ is the original image feature from the \textit{l}-th sample. Assuming that there are $C$ classes in this dataset, $o_i$ will be annotated with multi-label semantic $l_{l}=\left[l_{i 1}, \ldots, l_{i c}\right]$, where $l_{i j}=1$ indicates that $o_{l}$ belongs to the \textit{j}-th class, and $l_{i j}=0$  if not. The image-feature matrix is noted as $V$, and the label matrix as $L$ for all instances. The pairwise multi-label similarity matrix $S$ is used to describe semantic similarities between each of the two instances, where $S_{i,j}=1$  means that $O_i$ is semantically similar to $O_{j},$ otherwise $S_{i, j}=0$. In a multi-label setting, two instances $\left(O_{l}\right.$ and $\left.O_{j}\right)$ are annotated by multiple labels. Thus, we define $S_{i, j}=1$, if $O_{l}$ and $O_{j}$ share at least one label, otherwise $S_{i, j}=0$. The main goal in deep hashing retrieval is to identify a nonlinear hash function, i.e., $H:o \rightarrow h \in\{-1,1\}^{K}$, where $K$ is the length of each hash codes, to encode each item $o_{l}$ into a $K$-bit hash code $H_{i} \in\{-1,1\}$, whereby the correlation of all item pairs are maintained. The similarity between a hash code pair $H_{i}, H_{j}$ are evaluated by their Hamming distance $d i s_{H}\left(H_{i}, H_{j}\right),$ which might be a challenging and costly calculation \cite{61}. The inner-product $\left\langle H_{i}, H_{j}\right\rangle$ can be used as a surrogate which relates to hamming distance as follows:
\begin{equation}\label{eq1}
  dis_{H}=\dfrac{1}{2}\left(K-\left\langle H_{i}, H_{j}\right\rangle\right).
\end{equation}

\subsection{Self-supervised semantic network}
To enrich the semantic information in generated hash codes, we designed a self-supervised MLP network namely Semantic-Network to leverage abundant semantic correlations from multi-label annotations, the semantic information preserved by Semantic-Network will be further refined to perform as the guidance of the hash function learning process of Image-Network

Semantic-Network extracts high-dimensional semantic features thr-ough fully-connected layers with multi-label annotations as inputs $\left( \text{i.e.}, H_{i}^{l}=f^{l}\left(l_{l}, \theta^{l}\right)\right)$, where $f^l$ is the nonlinear hash function for Semantic-Network, while $\theta^l$ denotes the parameters for Semantic-Network. With a sign function the learned $H^l$ can be discretized into binary codes:
\vspace{-5pt}
\begin{equation}\label{eq2}
  B^{l}=\operatorname{sign}\left(H^{l}\right) \in\{-1,1\}^{K}.
\end{equation}

\vspace{-5pt}
For comprehensive preservation of semantic information especially in multi-label scenarios, the abstract semantic features $F^{l}=\left[F_{l}^{l}, \ldots, F_{n}^{l}\right]$ of Semantic-Network are also exploited to supervise the semantic learning of Image-Network.

\subsubsection{Cosine-distance-based similarity evaluation}
In Hamming space, the similarity of two hash codes $H_i$, $H_j$ can be defined by the Hamming distance dist$_H(*,*)$. To preserve the similarity of item pairs, whereby similar pairs are clustered and dissimilar pairs scattered, a similarity loss function of Semantic-Network is defined as follows:
\begin{equation}\label{eq3}
  \begin{aligned}
    J_{s}=&\sum\limits_{i, j=1}^{n}\left(s_{i, j} \operatorname{dis}_{H}\left(H_{i}, H_{j}\right)\right.\\ &\left.+\left(1-s_{i, j}\right) \max \left(m-\operatorname{dis}_{H}\left(H_{i}, H_{j}\right), 0\right)\right)
  \end{aligned}
\end{equation}
Where $J_s$ denotes the similarity loss function, by which the similarity of two generated hash codes $H_i$ and $H_j$ can be preserved. $\mathrm{dis}_H(H_i, H_j)$ represents the Hamming distance between $H_i$ and $H_j$. To avoid the collapsed scenario\cite{47}, a contrastive form of loss function is applied with a margin parameter $m$, with which the hamming distance of generated hash code pairs are expected to be less than $m$. With the mentioned relationship (\ref{eq1}) between Hamming distance and inner-product, the similarity loss can be redefined as:
\begin{equation}\label{eq4}
  \begin{aligned}
    J_{s}=&\sum\limits_{i, j=1}^{n} \dfrac{1}{2}\left(s_{i, j} \max \left(m-\left\langle H_{i}, H_{j}\right\rangle, 0\right)\right.\\ &\left.+\left(1-s_{i, j}\right) \max \left(m+\left\langle H_{i}, H_{j}\right\rangle, 0\right)\right)
  \end{aligned}
\end{equation}
Where the margin parameter induce the inner-product of dissimilar pairs to be less than $-m$, while that of similar ones to be larger than $m$, note that this form of contrastive similarity constraint derives from\cite{84} where margin is a hyper-parameter which is different from the hyper-plane margin used in SVM-based methods\cite{85,86}. For enhancement of similarity preservation, we expect the similarity constraint to be extended by ensuring the discrimination of deep semantic features. However because of the difference between the distributions of features from Semantic-Network and Image-Network, the inner-product $\langle . , .\rangle \in(-\infty, \infty)$ will no longer be a plausible choice for the similarity evaluation between the semantic features of the two networks. As the choice of margin parameter $m$ is ambiguous. One way to resolve this flaw is to equip the two networks with the same activate function, for example a sigmoid or tanh, at the output of the semantic layer to limit the scale of output features to a fixed range, nevertheless we expect both of the networks to maintain their own scale of feature representations. Considering the fact that hash codes are discretized to either -1 or 1 at each bit, meanwhile all generated hash codes have the same length $K$, therefore in the similarity evaluation in Hamming space, we choose to focus more on the angles between hash codes, instead of the absolute distance between them. Hence we adopt the cosine distance $\cos (. , .)$ as a replacement:
\begin{equation}\label{eq5}
  \cos \left(H_{i}, H_{j}\right)=\dfrac{<H_{i}, H_{j}>}{\|H_{i}\|\|H_{j}\|}
\end{equation}
Where $\cos \left(H_{i}, H_{j}\right) \in(-1,1)$. Although pairwise label information is adopted to store the semantic similarity
of hash codes, the label information is not fully exploit. Thus Semantic-Network will further exploit semantic information with an auxiliary classifier as shown in Fig. \ref{fig_1}. Many recent works directly map the learned binary codes into classification predictions by using a linear classifier\cite{28,29}. To prevent the interference between the classification stream and hashing stream, and to avoid the classification performance being too sensitive to the length of hash codes, we jointly learn the classification task and hashing task under a multi-task learning scheme without mutual interference\cite{35,36}.

The final object function of Semantic-Network can be formulated as:
\vspace{-10pt}
{\small{
\begin{equation}\label{eq6}
  \begin{aligned}
  &\min _{B^{l}, \theta^{l}, \hat{L}^{l}} J_{L a b} \\
  = &\alpha J_{1}+\lambda J_{2}+\eta J_{3}+\beta J_{4} \\
  = &\alpha \sum\limits_{i, j=1}^{n} \frac{1}{2} S_{i, j} \max \left(m-\Delta_{i, j}^{l}, 0\right)+\frac{1}{2}\left(1-S_{i, j}\right) \max \left(m+\Delta_{i, j}^{l}, 0\right) \\
  +&\lambda \sum\limits_{i, j=1}^{n} \frac{1}{2} S_{i, j} \max \left(m-\Gamma_{i, j}^{l}, 0\right)+\frac{1}{2}\left(1-S_{i, j}\right) \max \left(m+\Gamma_{i, j}^{l}, 0\right) \\
  +&\eta\left\|\hat{L}^{l}-L\right\|_{2}^{2} +\beta\left\|H^{l}-B^{l}\right\|_{2}^{2}
  \end{aligned}
\end{equation}}}
Where the margin is a manually-selected hyper-parameter $m\in(0,1)$. Taking semantic labels as inputs and being trained in self-supervised manner, it’s relatively easy for Se- mantic-Network to achieve robust retrieval accuracy, and it’s performance is not sensitive to the selection of margin value, with respect to the sensitivity analysis latter in \ref{sec_432}., it can consistently achieve robust performance when $m$ is relatively small, so we directly set it as 0 in experiments. $J_{l}$ and $J_{2}$ are the similarity loss for the learned semantic features and hash codes respectively with $\Delta_{i, j}^{l}=\cos \left(F_{i}^{l}, F_{j}^{l}\right), \Gamma_{i, j}^{l}=\cos \left(H_{i}^{l}, H_{j}^{l}\right) .$ The classification loss $J_{3}$ calculates the difference between input labels and predicted labels. $J_{4}$ is the quantization loss for the discretization of learned hash codes.

\subsubsection{Asymmetric guidance mechanism}
In existing self-supervised hashing methods \cite{29, 62}, the self-super-vised network normally guides the deep hashing network with a symmetric point-to-point strategy, hash codes generated by one mini-batch of image are directly associated with the hash codes generated by the corresponding mini-batch of labels. Under such mechanism, the global semantic information is insufficiently transferred to deep hashing network, meanwhile the similarity search process excessively focus on the semantics that frequently appear, whereas the semantics with lower frequency of occurrence are relatively neglected. In this paper, we motivated to alleviate the mentioned drawbacks of existing guidance mechanism. Inspired by asymmetric hashing methods, where the asymmetric association between instances have significantly empowered the effectiveness of similarity search. As illustrated in Fig. \ref{fig_1}, we train Semantic-Network until convergence, and refine the semantic information preserved by it from the entire training set labels, this is achieved by using Semantic-Network to generate binary code and semantic features for deduplicated multi-label annotations of the entire training set(i.e., each case of multi-label annotation is taken as input for only once),  the generated binary codes constitute a semantic code dictionary $U=\left\{u_{i}\right\}_{i=1}^{C}$ where $u_{i} \in[-1,1]$ and a corresponding semantic feature dictionary $Q=\left\{q_{i}\right\}_{i=1}^{C}$, where $C$ is the total number of deduplicated training set labels, both semantic dictionaries can be addressed by multi-label annotations.

\subsection{Deep feature learning network}
We apply an end-to-end convolutional neural network namely Image-Network for image feature learning, which can extract and embed deep visual features from images into high dimensional semantic features and simultaneously proj- ect them into output representations for multi-label classification task and hashing task, similar to Semantic-Network, two tasks are learned simultaneously under a multi-task learning framework. The semantic feature extraction and hash function learning of Image-Network will be supervised by the semantic maps $U$ and $Q$ generated in Semantic-Network using an asymmetric learning strategy, the asymmetric similarity constraint can be formulated as follows:
\begin{equation}\label{eq7}
  \begin{aligned}
    J_{s}=\sum\limits_{i=1}^{n} \sum\limits_{j=1}^{c} \frac{1}{2}&\left(s_{i, j} \max \left(m-\cos \left(H_{i}, u_{j}\right), 0\right)\right.\\ &\left.+\left(1-s_{i, j}\right) \max \left(m+\cos \left(H_{i}, u_{j}\right), 0\right)\right)
  \end{aligned}
\end{equation}
where $s_{i,j}$ is an asymmetric affinity matrix.

\subsubsection{Margin-scalable constraint}
In most contrastive or triplet similarity constraints used in deep hash methods\cite{38,39,32}, the choice of the margin parameter mainly relies on manual tuning. As demonstrated in \ref{sec_432}, we observe that, in comparison with the self-supervised Semantic-Network, the deep Image-Network is fairly sensitive to the choice of margin, which means that a good selection of margin is valuable for robust hash function learning. Additionally, in multi-label scenarios, it would be more desirable if the margin can be scaled to be larger for item pairs that share more semantic similarities than those less semantically similar pairs, in this case the scale of margin can be equivalent to the strength of constraint. Thus setting a single fixed margin value may downgrade the storage of similarity information. Holding the motivation of dynamically selecting optimized margin for each sampled instance pairs with respect to their exact degree of semantic similarity, we propose a margin-scalable similarity constraint based on the semantic maps generated by Semantic-Network. Relying on the insensitivity of Semantic-Network to selection of margin, we leverage information in semantic dictionaries to calculate scalable margin and to indicate relative semantic similarity, i.e., for two hash codes $H_{i}^{v}$ and $H_{j}^{v}$ generated by Image-Network, a pair of corresponding binary codes $q_{H_{i}^{v}}$ and $q_{H_{j}^{v}}$ are represented by addressing the semantic code map $Q$ with their semantic labels as index. The scalable margin $M_{H_{i}, H_{j}}$ for $H_{i}^{v}$ and $H_{j}^{v}$ is calculated by:
\begin{equation}\label{eq8}
  M_{H_{i}, H_{j}}=\max \left(0, \cos \left(q_{H_{i}^{v}}, q_{H_{j}^{v}}\right)\right)
\end{equation}
As $\cos \left(q_{H_{i}^{v}}, q_{H_{j}^{v}}\right) \in(-1,1)$, a positive cosine distance between item pairs in the semantic code dictionary will be assigned to similar item pairs and will be used by Image-Network to calculate their scalable margin, while the negative cosine distances will scale the margin to 0. This is due to the nature of multi-label tasks, where the ‘dissimilar' situation only refers to item pairs with none identical label. While for a similar item pair, the number of shared labels may come from a wide range. Thus in similarity preservation, dissimilar items are given a weaker constraint, whereas the similar pairs are constrained in a more precise and strict way.
For two sampled sets of hash codes or semantic features $G_1$ and $G_2$ with size of $n_1$ and $n_2$, the margin-scalable constraint $J_{ms}$ can be given by:

\begin{equation}\label{eq9}
  \begin{aligned}
  &J_{ms}\left( {{G_1},{G_2}} \right)\\
  &= \sum\limits_{i = 1}^{{n_1}} {\sum\limits_{j = 1}^{{n_2}} {\frac{1}{2}\left( {S_{i,j}}max\left( {{M_{{G_i},{G_j}}} - \cos \left( {G_1^i,G_2^j} \right),0} \right) \right.} } \\
  &+ \left( {1 - {S_{i,j}}} \right)max\left( {{M_{{G_i},{G_j}}} - \cos \left( {G_1^i,G_2^j} \right),0} \right)
  \end{aligned}
\end{equation}

The final object function of Image-Network can be formulated as:
\begin{equation}\label{eq10}
  \begin{aligned}
  &\min\limits_{B^{v}, \theta^{v}, \hat{L}^{v}} J_{\mathrm{Img}} \\
  =&\alpha J_{1}+\lambda J_{2}+\gamma J_{3} + \mu J_{4}+\eta J_{5}+\beta J_{6} \\
  =&\alpha {J_{ms}}\left( {{F^v},{F^v}} \right) + \lambda {J_{ms}}\left( {{H^v},{H^v}} \right)
  +\gamma {J_{ms}}\left( {{F^v},Q} \right)\\ 
  +&\mu {J_{ms}}\left( {{H^v},U} \right)
  +\eta \parallel {{\hat L}^v} - L\parallel _2^2 + \beta \parallel {H^v} - {B^v}\parallel _2^2
  \end{aligned}
\end{equation}
where $J_{l}$ and $J_{2}$ are margin-scalable losses for semantic features and hash codes generated by Image-Network, with symmetric association between instance pairs. $J_{3}$ and $J_{4}$ are margin-scalable losses with asymmetric guidance of semantic dictionaries $U$ and $Q$ on hash codes and semantic Features generated by Image-Network.  $J_{5}$ and $J_{6}$ are classification loss and quantization loss similarly defined in Semantic-Network.

\subsection{Optimization}
It is noteworthy to mention that, the Image-Network is trained after the convergence of Semantic-Network is obtained. First we iteratively optimize the objective function (\ref{eq6}) by exploring multi-label information to learn $\theta^{l}, H^{l}$ and $\hat{L}^{l}$. With the finally trained Semantic-Network we obtain $U$ and $Q$. Then the parameters of Semantic-Network will be fixed, and $L_{\text {img}}$ wil be optimized through $\theta^{v}, H^{v}$ and $\hat{L}^{v}$ with the guidance of $U$ and $Q$. Finally, we obtain binary hash codes $B=\operatorname{sign}\left(H^{v}\right)$. The entire learning algorithm is summarized in Algorithm \ref{alg_1} in more detail.

\subsubsection{Optimization of Semantic-Network}
The gradient of $J_{L a b}$ w.r.t each Hash code $H_{i}^{l}$ in sampled mini-batch is
\begin{equation}\label{eq11}
\begin{array}{l}
\dfrac{\partial J_{L a b}}{\partial H_{i}^{l}}=\\
\left\{\begin{array}{l} 
  \sum\limits_{\substack{j=1 \\ s_{i, j}=1}}^{n} \dfrac{\lambda}{2}\left(m-\dfrac{H_{j}^{l}}{\| H_{i}^{l} \Vert \| H_{j}^{l} \Vert }+\dfrac{H_{i}^{l} \Gamma_{i, j}^{l}}{\| H_{i}^{l} \Vert_{2}^{2}}\right)+2 \beta\left(H_{i}^{l}-B_{i}^{l}\right)\\ 
  \hspace{25pt} \text { if }\ s_{i, j}=1 \text { and } \Gamma_{i, j}^{l}<m\\ 
  \quad \\
  \sum\limits_{\substack{j=1 \\ s_{i, j}=0}}^{n} \dfrac{\lambda}{2}\left(m+\dfrac{H_{j}^{l}}{\| H_{i}^{l} \Vert \| H_{j}^{l} \Vert }-\dfrac{H_{i}^{l} \Gamma_{i, j}^{l}}{\| H_{i}^{l} \Vert_{2}^2}\right)+2 \beta\left(H_{i}^{l}-B_{i}^{l}\right)\\ 
  \hspace{25pt} \text { if }\ s_{i, j}=0 \text { and } \Gamma_{i, j}^{l}>-m \end{array}\right.
\end{array}
\end{equation}
Where $\Gamma_{i, j}^{l} = \cos \left( {H_i^l,H_j^l} \right)$. $\dfrac{\partial J_{L a b}}{\partial F_{j}^{l}}$ can be obtained similarly, $\dfrac{\partial J_{L a b}}{\partial \theta^{l}}$ can be computed by using the chain rule, then $\theta^{l}$ can be updated for each iteration using Adam with back propagation.

\subsubsection{Optimization of Image-Network}
The gradient of $J_{\mathrm{Img}}$ w.r.t each Hash code $H_{i}^{v}$ in sampled mini-batch is
\begin{equation}\label{eq12}
\dfrac{\partial J_{\mathrm{Img}}}{\partial H_{i}^{v}}=\lambda \dfrac{\partial J_{2}}{\partial H_{i}^{v}}+\mu \dfrac{\partial J_{4}}{\partial H_{i}^{v}}+\beta \dfrac{\partial J_{6}}{\partial H_{i}^{v}}
\end{equation}
Where
$$
\begin{array}{l}
\dfrac{\partial J_{2}}{\partial H_{i}^{v}}=
\left\{\begin{array}{l} 
  \sum\limits_{\substack{j=1 \\ s_{i, j}=1}}^{n} \dfrac{1}{2}\left(M_{H_{i}, H_{j}}-\dfrac{H_{j}^{v}}{\| H_{i}^{v} \Vert \| H_{j}^{v} \Vert }+\dfrac{H_{i}^{v} \Gamma_{i, j}^{v}}{\| H_{i}^{v} \Vert _{2}^{2}}\right)\\ \hspace{25pt}\text { if }\ s_{i, j}=1 \text { and } M_{H_{i}, H_j}>\Gamma_{i, j}^{v}\\ 
  \quad \\
  \sum\limits_{\substack{j=1 \\ s_{i, j}=0}}^{n} \dfrac{1}{2}\left(\dfrac{H_{i}^{v} \Gamma_{i, j}^{v}}{\| H_{i}^{v} \Vert _{2}^2}-\dfrac{H_{j}^{v}}{\| H_{i}^{v} \Vert  \| H_{j}^{v} \Vert }-M_{H_{i}, H_{j}}\right)\\  \hspace{25pt}\text { if }\ s_{i, j}=0 \text { and } M_{H_{i}, H_{j}}>\Gamma_{i, j}^{v}
  \end{array}\right.
\end{array}
$$
Where $\Gamma_{i, j}^{v} = \cos \left( {H_i^v,H_j^v} \right)$. $\dfrac{\partial J_{6}}{\partial H_{i}^{v}}=2\left(H_{i}^{l}-B_{i}^{l}\right)$, the calculation of $\dfrac{\partial J_4}{\partial H_i^v}$ resembles $\dfrac{\partial J_2}{\partial H_i^v}$, $\dfrac{\partial J_{\mathrm{Img}}}{\partial F_{i}^{v}}$ can be obtained similarly to $\dfrac{\partial J_{\mathrm{Img}}}{\partial H_{i}^{v}}$, $\dfrac{\partial J_{\mathrm{Img}}}{\partial \theta^{v}}$ can be computed by using the chain rule, then $\theta^{v}$ can be updated for each iteration using SGD with back propagation.
\begin{algorithm}
  \caption{The learning algorithm of our \textit{SADH}}
  \label{alg_1}
  \begin{algorithmic}
    \Require 
    \State Image set $V$, Label set $L$
    \Ensure 
    \State semantic feature map $Q$, and semantic code map $U$, parameters $\theta^{v}$ for Image-Network, 
    \State Optimal code matrix for Image-Network $B^{v}$
    \Require 
    \State Initialize network parameters $\theta^{l}$ and $\theta^{v}$
    \State Hyper-parameters: $\alpha, \lambda, \gamma, \mu, \eta, \beta, m$
    \State Mini-batch size $M$, learning rate: $lr$
    \State maximum iteration numbers $t^{l}, t^{v}$ 
    \State\hspace{-10pt}\textbf{Stage1:} Hash learning for the self-supervised network (Semantic-Network)
    \For{$t^l$ iteration}
    \State Calculate derivative using formula (\ref{eq11})
    \State Update $\theta^l$ by using Adam and back propagation
    \EndFor
    \State Update semantic feature map $Q$ and semantic code map $U$ by Semantic-Network for each semantic as input
    \State\hspace{-10pt}\textbf{Stage2:} Hash learning for the feature learning network (Image-Network)
    \For{$t^v$ iteration}
    \State Calculate derivative using formula (\ref{eq12})
    \State Update $\theta^v$ by using SGD and back propagation
    \EndFor
    \State Update the parameter $B^{v}$ by $B^{v}=\operatorname{sign}\left(H^{v}\right)$
  \end{algorithmic}
\end{algorithm}
\subsection{Extension to cross-modal hashing}
\label{sec_35}
As mentioned in \ref{sec_24}, hashing in Cross-modal scenarios has arouse extensive attention of many researchers, in which a common Hamming space is expected to be learned to perform mutual retrieval between data of heterogeneous modalities. In this paper, we mainly consider the single-modal retrieval of image data, but the flexibility of margin-scalable constraint and asymmetric guidance mechanism allows us to readily extend our SADH algorithm to achieve cross-modal hashing. Suppose the training instances consists of $N$ different modalities, with corresponding hash codes $H^{j}, j=1, \ldots, N$, and semantic features $F^{j}, j=1, \ldots, N$. Then the extension of our proposed method in Eq. (4) can be formulated as:
\begin{equation}\label{eq9}
  \begin{aligned}
  \min _{B^{j}, \theta^{j}, \hat{L}^{j}} \sum\limits_{j = 1}^N {\alpha {J_{ms}}\left( {{F^{\rm{j}}},{F^j}} \right) + \lambda {J_{ms}}\left( {{H^j},{H^j}} \right)} \\
 + \gamma {J_{ms}}\left( {{F^j},Q} \right) + \mu {J_{ms}}\left( {{H^j},U} \right)\\
 + \eta \parallel {{\hat L}^j} - {L^j}\parallel _2^2 + \beta \parallel {H^j} - {B^j}\parallel _2^2
  \end{aligned}
\end{equation}
Without loss of generality, following methods like \cite{97, 98, 99, 89}, we focus on cross-modal retrieval for bi-modal data (i.e., image and text) in experimental analysis.

\section{Experiments and analysis}
In this section, we conducted extensive experiments to verify three main issues of our proposed SADH method: (1) To illustrate the retrieval performance of SADH compared to existing state-of-the-art methods. (2) To evaluate the improvements of efficiency in our method compared to other methods. (3) To verify the effectiveness of different modules proposed in our method.

\subsection{Datasets and experimental settings}
The evaluation is based on four mainstream image retrieval datasets: CIFAR-10\cite{40}, NUS-WIDE\cite{15}, MIRFlickr-25K\cite{16}, MS-COCO\cite{102}.

\textbf{CIFAR-10:} CIFAR-10 contains 60,000 images with a resolution of $32\times 32$. These images are divided into 10 different categories, each with 6,000 images. In the CIFAR-10 experiments, following\cite{41}, we select 100 images per category as testing set(a total of 1000) and query set, the remaining as database(a total of 59000), 500 images per category are selected from the database as a training set(a total of 5000).

\textbf{NUS-WIDE:} NUS-WIDE contains 269,648 image-text pairs. This data set is a multi-label image set with 81 ground truth concepts. Following a similar protocol as in \cite{28,41}, we use the subset of 195,834 images which are annotated by the 21 most frequent classes (each category contains at least 5,000 images). Among them, 100 image-text pairs and 500 image-text pairs are randomly selected in each class as the query set (2100 in total) and the training set (10500 in total), respectively. The remaining 193734 image-text pairs are selected as database. 

\textbf{MIRFlickr-25K:} The MIRFlickr25K dataset consists of 25,000 images collected from the Flickr website. Each instance is annotated by one or more labels selected from 38 categories. We randomly selected 1,000 images for the query set, 4,000 images for the training set and the remaining images as the retrieval database.

\textbf{MS-COCO:} The MS-COCO dataset consists of 82,783 training images and 40,504 validation images, each image is annotated with at least one of the 80 semantics, we combine the training set and validation set and prune the images with no categories, which gives us 122,218 images. For cross-modal retrieval, the text instances are presented in form of 2028-dimensional Bag-of-Word vectors.

For image retrieval, we compare our proposed SADH with several state-of-the-art approaches including LSH \cite{1}, SH \cite{8}, ITQ \cite{2}, LFH \cite{43}, DSDH \cite{28}, HashNet \cite{12}, DPSH \cite{9}, DBDH \cite{44}, CSQ \cite{63} and DSEH \cite{29} on all the four datasets. For cross-modal retrieval, we compare our SADH with 3 state-of-the-are deep cross-modal hashing frameworks including DCMH\cite{97}, PRDH\cite{98}, SSAH\cite{62}. These methods are briefly introduced as follows:

1. Locality-Sensitive Hashing (LSH) \cite{1} is a data-inde-pendent hashing method that employs random projections as hash function.

2. Spectral Hashing (SH) \cite{8}is a spectral method which transfers the original problem of finding the best hash codes for a given dataset into the task of graph partitioning.

3. Iterative quantization (ITQ) \cite{2} is a classical unsupervised hashing method. It projects data points into a low dimensional space by using principal component analysis (PCA), then minimize the quantization error for hash code learning.

4. Latent Factor Hashing (LFH) \cite{43} is a supervised method based on latent hashing models with convergence guarantee and linear-time variant.

5. Deep Supervised Discrete Hashing (DSDH) \cite{28} is the first supervised deep hashing method that simultaneously utilize both semantic labels and pairwise supervised information, the hash layer in DSDH is constrained to be binary codes.

6. HashNet \cite{12} is a supervised deep architecture for hash code learning, which includes a smooth activation function to resolve the ill-posed gradient problem during training.

7. Deep pairwise-supervised hashing (DPSH) \cite{9} is a representative deep supervised hashing method that jointly performs feature learning and hash code learning for pairwise application.

8. Deep balanced discrete hashing for image retrieval (DBDH) \cite{44} is a recent supervised deep hashing method which uses a straight-through estimator to actualize discrete gradient propagation.

9. Central Similarity Quantization for Efficient Image and Video Retrieval (CSQ) \cite{63} defines the correlation of hash codes through a global similarity metric, to identify a common center for each hash code pairs.

10. Deep Joint Semantic-Embedding Hashing (DSEH) \cite{29} is a supervised deep hashing method that employs a self-supervised network to capture abundant semantic information as guidance of a feature learning network.

11. Deep cross modal hashing (DCMH)\cite{97} is a supervised deep hashing method that integrates feature learning and hash code learning in an end-to-end framework.

12. Pairwise Relationship Guided Deep Hashing (PRDH) \cite{98} is a supervised deep hashing method that utilize both intra-modal and inter-modal pairwise constraints to search for similarity information.

13. Self-supervised adversarial hashing networks for cro- ss-modal retrieval(SSAH) \cite{62} is a deep supervised cross-modal method that utilize a self-supervised network to constitute a common semantic space to bridge data from image modality and text modality.

Among the above approaches, LSH\cite{1}, SH\cite{8}, ITQ\cite{2}, LFH \cite{43} are non-deep hashing methods, for these methods, 4096-dimentional deep features extracted from Alexnet \cite{23} are utilized for two datasets: NUS-WIDE and CIFAR-10 as inputs. The other six baselines (i.e., DSDH, HashNet, DPSH, DBDH and DSEH) are deep hashing methods, for which images on three dataset (i.e., NUS-WIDE, CIFAR-10 and MIRFlickr-25k) are resized to $224\times 224$ and used as inputs. LSH, SH, ITQ, LFH, DSDH, HashNet, DPSH, DCMH and SSAH are carefully carried out based on the source codes provided by the authors, while for the rest of the methods, they are carefully implemented by ourselves using parameters as suggested in the original papers.

We evaluate the retrieval quality by three widely used evaluating metrics: Mean Average Precision (MAP), Pre-cision-Recall curve, and Precision curve with the number of top returned results as variable (topK-Precision).

Specifically, given a query instance q, the Average Precision (AP) is given by:
$$
A P(q)=\dfrac{1}{n_{q}} \sum\limits_{i=1}^{n_{database}} P_{q_{i}} I(i)
$$
Where $n_{database}$ is the total number of instances in the databa-se, $n_q$ is the number of similar samples, $P_{q_{i}}$ is the probability of instances of retrieval results being similar to the query instance at cut-off $i$, And $I(i)$ is the indicator function that indicates the \textit{i}-th retrieval instance $I$ is similar to query image to $q$, if $I(i)=1$, and $I(i)=0$ otherwise.

The larger the MAP is, the better the retrieval performance. Since NUS-WIDE is relatively large, we only consider the top 5,000 neighbors (MAP@5000), when computing MAP for NUS-WIDE, while for CIFAR-10 and MIRFlic-kr-25K, we calculate MAP for the entire retrieval database (MAP@ALL).

\subsection{Implementation details}
Semantic-Network is built with four fully-connected layers, with which the input labels are transformed into hash codes $(L \rightarrow 4096 \rightarrow 2048 \rightarrow N)$. Here the output includes both the $K$-dimensional hash code and the $C$-dimensional multi-label predictions, $N=K+C$.

We built ImageNet based on Resnet50, the extracted visual features of Resnet are embedded into 2048-dimensional semantic features, which is followed by the two extra layers (i.e., Hash layer and Classification layer) with $K$ nodes for hash code generation and $C$ nodes for classification. It is noted that except for output layers, the network is pre-trained on ImageNet dataset.
The implementation of our method is based on the Pytorch framework and executed on NVIDIA TITAN X GPUs for 120 epochs of training. The hyper-parameters in Semantic-Network, set $\alpha, \lambda, \eta$, $\beta$ are set to 2,0.5,0.5,0.1 respectively. The hyper-parameters in Image-Network, $\alpha, \lambda, \gamma, \mu, \eta, \beta$ to 0.01,1,0.01,1,2 and 0.05 respectively. As can be 
observed from Fig. \ref{fig_8}, Semantic-Network maintains a stable and robust retrieval performance under different choices of margin parameter, especially for small margin parameters. Hence we simply set m to 0 for all the scenarios.

The Adam optimizer\cite{45} is applied to Semantic-Network, while the stochastic Gradient descent (SGD) method is applied to Image-Network. The batch size is set to 64. The learning rates are chosen from $10^{-3}$ to $10^{-8}$ with a momentum of 0.9.

\subsection{Performance evaluation}
\subsubsection{Comparison to State of the Art}
To validate the retrieval performance of our method for image retrieval, we compare the experimental results of SAD-H with other state-of-the-art methods including LSH \cite{1}, SH \cite{8}, ITQ \cite{2}, LFH \cite{43}, DSDH \cite{28}, HashNet \cite{12}, DPSH \cite{9}, DBDH \cite{44}, CSQ \cite{63} and DSEH \cite{29} on CIFAR-10, NUS-WIDE, MIRFlickr-25K and MS-COCO. Table \ref{tab_1} shows the top 10 retrieved images in database for 3 sampled images in MIRFlickr-25K, it can be observed that in difficult cases, SADH reveals better semantic consistency than HashNet. Table \ref{tab_2} to Table \ref{tab_5} report the MAP results of different methods, note that for NUS-WIDE, MAP is calculated for the top 5000 returned neighbors. Fig. \ref{fig_2}-\ref{fig_7} show the overall retrieval performance of SADH compared to other baselines in terms of precision-recall curve and precision curves by varying the number of top returned images, shown from 1 to 1000, on NUS-WIDE, CIFAR-10, MS-COCO and MIRFlickr-25K respectively. SADH substantially outperforms all other state-of-the-art methods. It can be noticed that SADH outperforms other methods for almost all the leng-ths of hash bits with a steady performance on both datasets. This is due to the multi-task learning structure in our method with which the classification output and hashing output are obtained independently, and the two tasks are not mutually interfered. It is also noteworthy that, with abundant semantic information leveraged from the self-supervised network and the pairwise information derived from the margin-scalable constraint, SADH obtained an impressive retrieval perfor-mance on both single-label datasets and multi-label datasets.  
\begin{strip}

  \quad

  \quad

  \includegraphics[width=\textwidth]{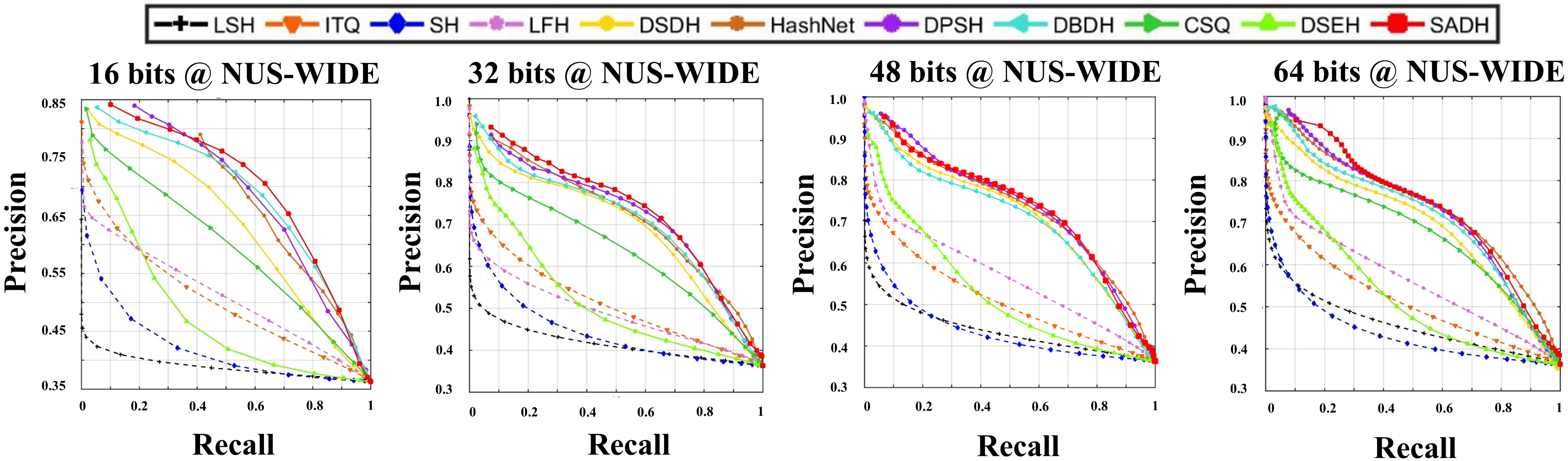}
  \captionof{figure}{precision-recall curves on NUS-WIDE.}
  \label{fig_2}

  \includegraphics[width=\textwidth]{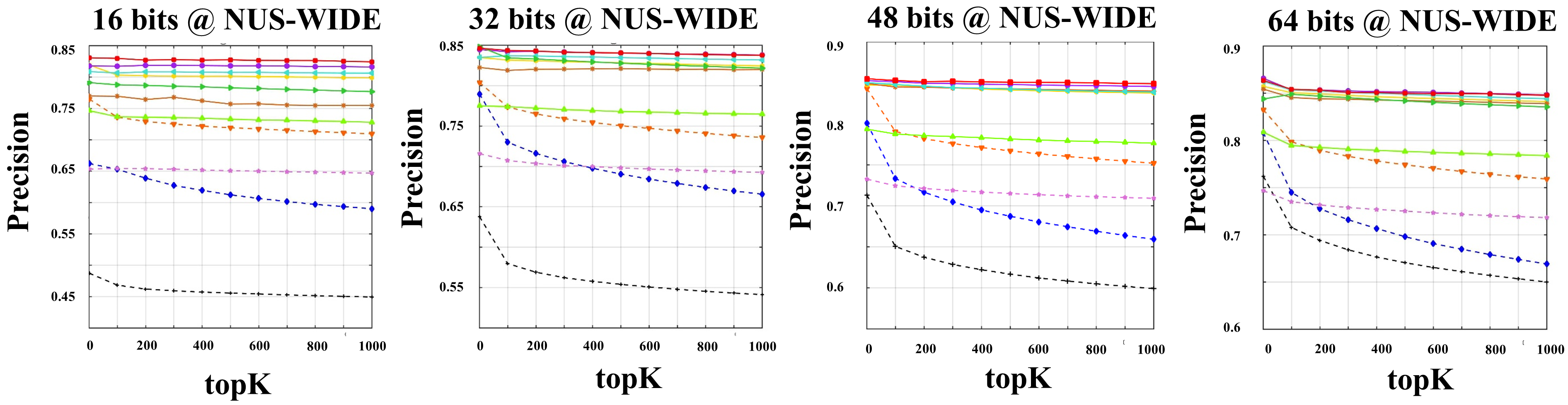}
  \captionof{figure}{TopK-precision curves on NUS-WIDE.}
  \label{fig_3}

  \quad

  \quad

  \includegraphics[width=\textwidth]{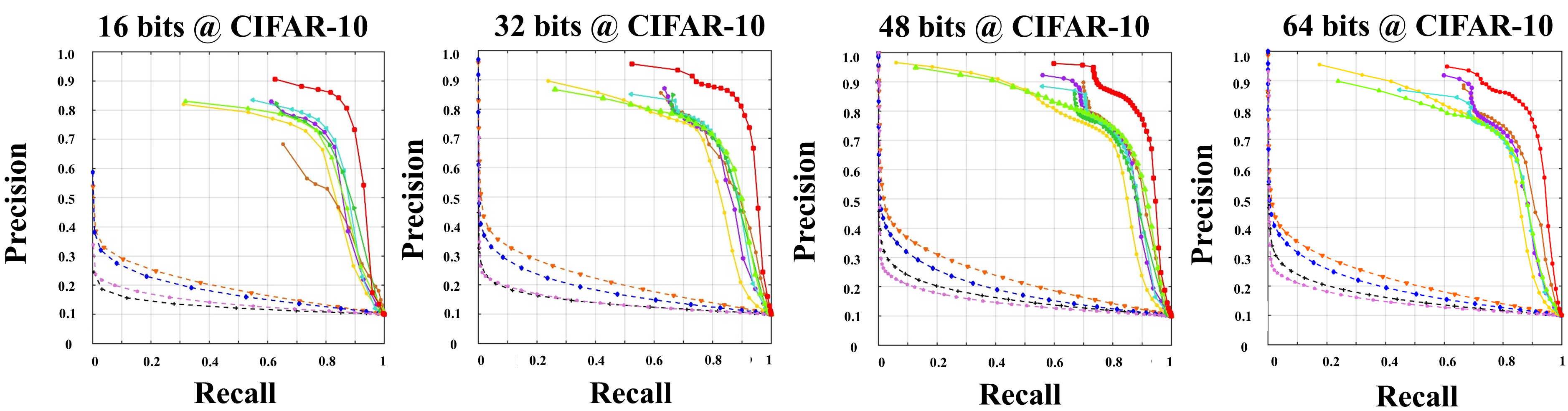}
  \captionof{figure}{precision-recall curves on CIFAR-10.}
  \label{fig_4}

  \quad

  \quad

  \includegraphics[width=\textwidth]{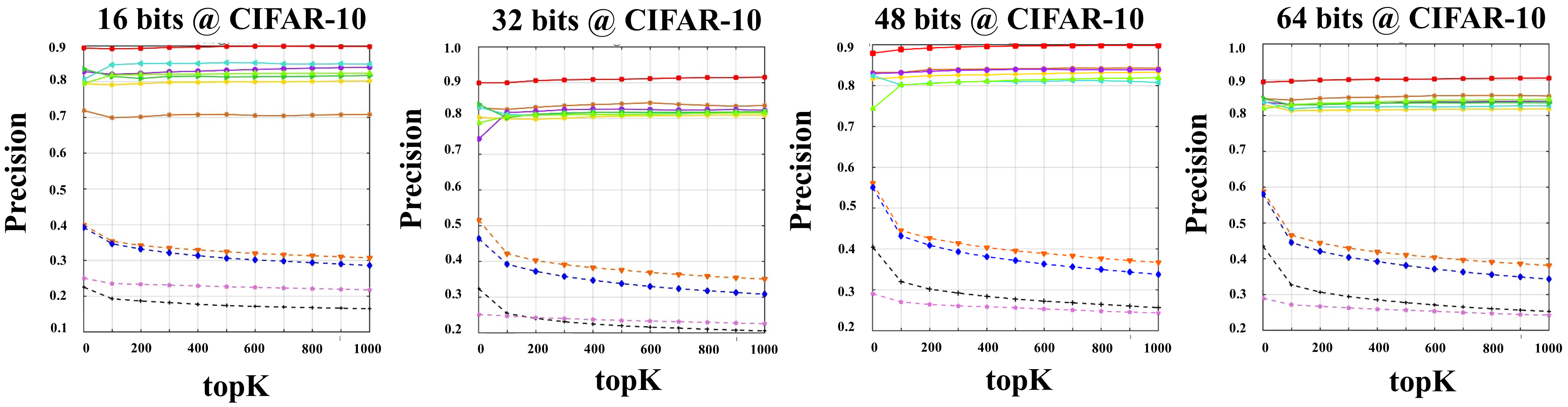}
  \captionof{figure}{TopK-precision curves curves on CIFAR-10.}
  \label{fig_5}

  \quad 

  \quad

  \includegraphics[width=\textwidth]{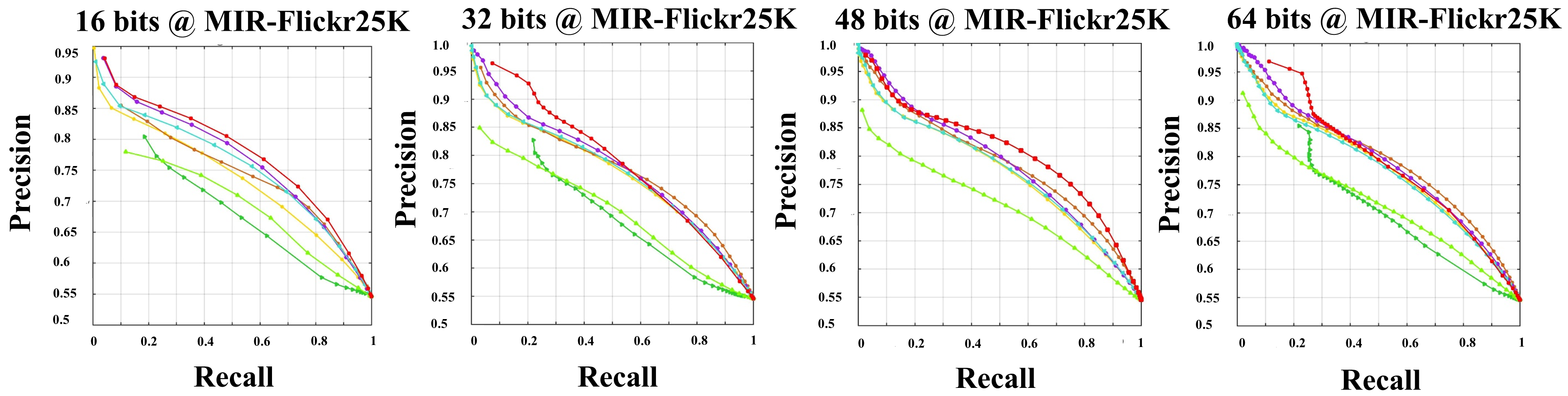}
  \captionof{figure}{precision-recall curves on MIR-Flickr25K.}
  \label{fig_6}

  \quad

  \quad

  \includegraphics[width=\textwidth]{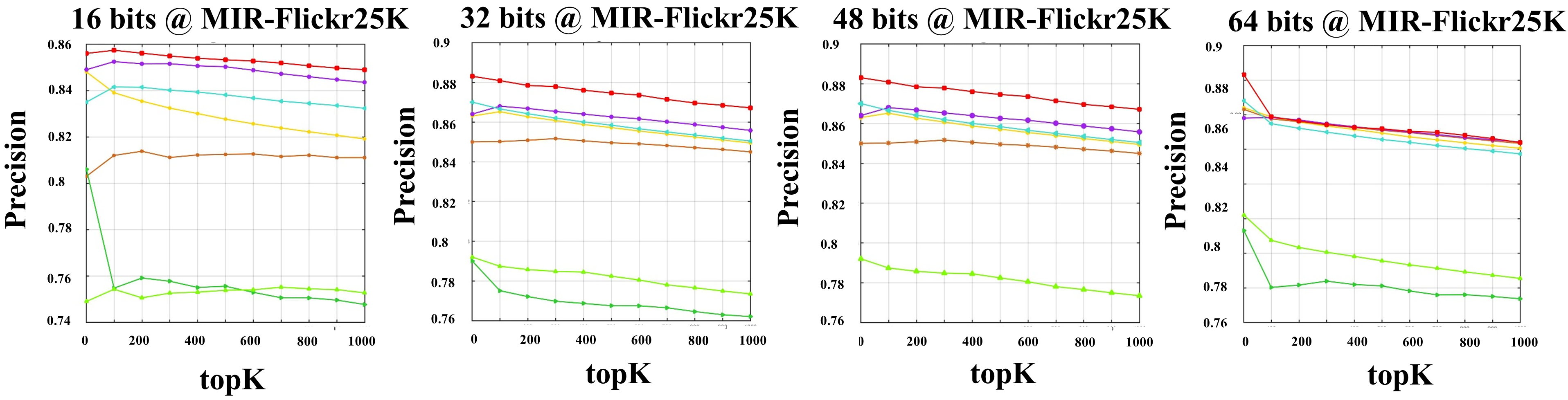}
  \captionof{figure}{TopK-precision curves on MIR-Flickr25K.}
  \label{fig_7}

    \centering
    \captionof{table}{Examples of top 10 retrieved images by SADH and DSDH on MIRFlickr-25K for 48 bits. The semantically incorrect  images are marked with a red border.}
    \label{tab_1}
    \resizebox{\textwidth}{!}{%
    \begin{tabular}{p{0.088\textwidth}<{\centering}p{0.088\textwidth}<{\centering}p{0.088\textwidth}<{\centering}p{0.088\textwidth}<{\centering}p{0.088\textwidth}<{\centering}p{0.088\textwidth}<{\centering}p{0.088\textwidth}<{\centering}p{0.088\textwidth}<{\centering}p{0.088\textwidth}<{\centering}p{0.088\textwidth}<{\centering}p{0.088\textwidth}<{\centering}} \\ 
    \toprule
    Query             & \multicolumn{10}{c}{Top10 Retrieved Images} \\ 
    \midrule
    \multirow{6}{*}{\begin{minipage}{0.088\textwidth}
      \vspace{30pt}
      \includegraphics[width=1.16\textwidth]{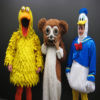}
      \centerline{Portrait}
      \centerline{Indoor}
      \centerline{people}
    \end{minipage}} & \multicolumn{10}{l}{SADH}                   \\
                      &  {\begin{minipage}{0.088\textwidth}
                        \includegraphics[width=1.16\textwidth]{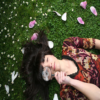}
                      \end{minipage}}  &  {\begin{minipage}{0.088\textwidth}
                        \includegraphics[width=1.16\textwidth]{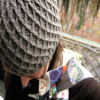}
                      \end{minipage}}  & {\begin{minipage}{0.088\textwidth}
                        \includegraphics[width=1.16\textwidth]{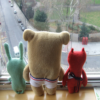}
                      \end{minipage}}   & {\begin{minipage}{0.088\textwidth}
                        \includegraphics[width=1.16\textwidth]{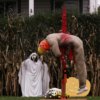}
                      \end{minipage}}   &  {\begin{minipage}{0.088\textwidth}
                        \includegraphics[width=1.16\textwidth]{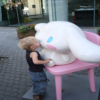}
                      \end{minipage}}  & {\begin{minipage}{0.088\textwidth}
                        \includegraphics[width=1.16\textwidth]{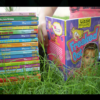}
                      \end{minipage}}  & {\begin{minipage}{0.088\textwidth}
                        \includegraphics[width=1.16\textwidth]{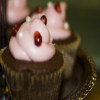}
                      \end{minipage}}  & ~{\begin{minipage}{0.088\textwidth}
                        \includegraphics[width=1.16\textwidth]{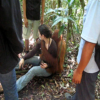}
                      \end{minipage}}  & {\begin{minipage}{0.088\textwidth}
                        \includegraphics[width=1.16\textwidth]{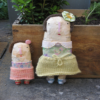}
                      \end{minipage}}  & {\begin{minipage}{0.088\textwidth}
                        \includegraphics[width=1.16\textwidth]{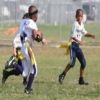}
                      \end{minipage}}  \\ 
                      &    &    &    &    &    &    &   &   &   & \\ \cline{2-11}
                      & \multicolumn{10}{l}{HashNet}                   \\
                      &  {\begin{minipage}{0.088\textwidth}
                        \includegraphics[width=1.16\textwidth]{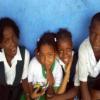}
                      \end{minipage}}  &  {\begin{minipage}{0.088\textwidth}
                        \includegraphics[width=1.16\textwidth]{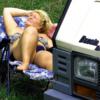}
                      \end{minipage}}  & {\begin{minipage}{0.088\textwidth}
                        \includegraphics[width=1.16\textwidth]{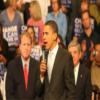}
                      \end{minipage}}   & {\begin{minipage}{0.088\textwidth}
                        \includegraphics[width=1.16\textwidth]{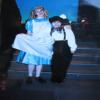}
                      \end{minipage}}   &  {\begin{minipage}{0.088\textwidth}
                        \includegraphics[width=1.16\textwidth]{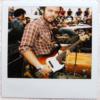}
                      \end{minipage}}  & {\begin{minipage}{0.088\textwidth}
                        \includegraphics[width=1.16\textwidth]{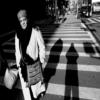}
                      \end{minipage}}  & {\begin{minipage}{0.088\textwidth}
                        \includegraphics[width=1.16\textwidth]{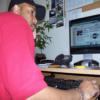}
                      \end{minipage}}  & ~{\begin{minipage}{0.088\textwidth}
                        \includegraphics[width=1.16\textwidth]{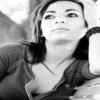}
                      \end{minipage}}  & {\begin{minipage}{0.088\textwidth}
                        \includegraphics[width=1.16\textwidth]{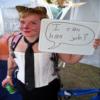}
                      \end{minipage}}  & {\begin{minipage}{0.088\textwidth}
                        \includegraphics[width=1.16\textwidth]{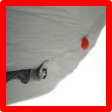}
                      \end{minipage}} \\
                      &    &    &    &    &    &    &   &   &   &   \\ 
    \midrule
    \multirow{6}{*}{\begin{minipage}{0.088\textwidth}
      \vspace{35pt}
      \includegraphics[width=1.16\textwidth]{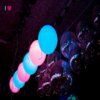}
      \centerline{Indoor}
      \centerline{Night}
    \end{minipage}} & \multicolumn{10}{l}{SADH}                   \\
    &  {\begin{minipage}{0.088\textwidth}
      \includegraphics[width=1.16\textwidth]{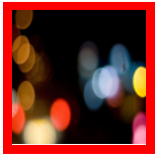}
    \end{minipage}}  &  {\begin{minipage}{0.088\textwidth}
      \includegraphics[width=1.16\textwidth]{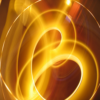}
    \end{minipage}}  & {\begin{minipage}{0.088\textwidth}
      \includegraphics[width=1.16\textwidth]{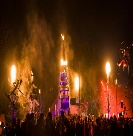}
    \end{minipage}}   & {\begin{minipage}{0.088\textwidth}
      \includegraphics[width=1.16\textwidth]{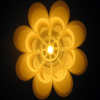}
    \end{minipage}}   &  {\begin{minipage}{0.088\textwidth}
      \includegraphics[width=1.16\textwidth]{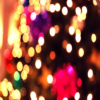}
    \end{minipage}}  & {\begin{minipage}{0.088\textwidth}
      \includegraphics[width=1.16\textwidth]{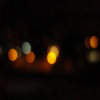}
    \end{minipage}}  & {\begin{minipage}{0.088\textwidth}
      \includegraphics[width=1.16\textwidth]{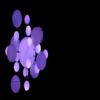}
    \end{minipage}}  & ~{\begin{minipage}{0.088\textwidth}
      \includegraphics[width=1.16\textwidth]{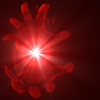}
    \end{minipage}}  & {\begin{minipage}{0.088\textwidth}
      \includegraphics[width=1.16\textwidth]{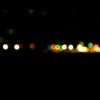}
    \end{minipage}}  & {\begin{minipage}{0.088\textwidth}
      \includegraphics[width=1.16\textwidth]{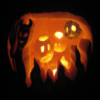}
    \end{minipage}}  \\ 
    &    &    &    &    &    &    &   &   &   & \\ \cline{2-11}
    & \multicolumn{10}{l}{HashNet}                   \\
    &  {\begin{minipage}{0.088\textwidth}
      \includegraphics[width=1.16\textwidth]{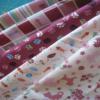}
    \end{minipage}}  &  {\begin{minipage}{0.088\textwidth}
      \includegraphics[width=1.16\textwidth]{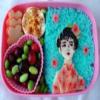}
    \end{minipage}}  & {\begin{minipage}{0.088\textwidth}
      \includegraphics[width=1.16\textwidth]{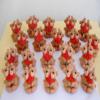}
    \end{minipage}}   & {\begin{minipage}{0.088\textwidth}
      \includegraphics[width=1.16\textwidth]{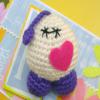}
    \end{minipage}}   &  {\begin{minipage}{0.088\textwidth}
      \includegraphics[width=1.16\textwidth]{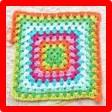}
    \end{minipage}}  & {\begin{minipage}{0.088\textwidth}
      \includegraphics[width=1.16\textwidth]{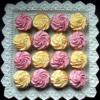}
    \end{minipage}}  & {\begin{minipage}{0.088\textwidth}
      \includegraphics[width=1.16\textwidth]{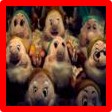}
    \end{minipage}}  & ~{\begin{minipage}{0.088\textwidth}
      \includegraphics[width=1.16\textwidth]{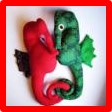}
    \end{minipage}}  & {\begin{minipage}{0.088\textwidth}
      \includegraphics[width=1.16\textwidth]{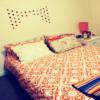}
    \end{minipage}}  & {\begin{minipage}{0.088\textwidth}
      \includegraphics[width=1.16\textwidth]{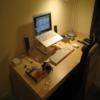}
    \end{minipage}} \\
    &    &    &    &    &    &    &   &   &   &   \\ 
    \midrule
    \multirow{6}{*}{\begin{minipage}{0.088\textwidth}
      \vspace{35pt}
      \includegraphics[width=1.16\textwidth]{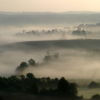}
      \centerline{Clouds}
      \centerline{sky}
    \end{minipage}} & \multicolumn{10}{l}{SADH}                   \\
    &  {\begin{minipage}{0.088\textwidth}
      \includegraphics[width=1.16\textwidth]{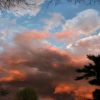}
    \end{minipage}}  &  {\begin{minipage}{0.088\textwidth}
      \includegraphics[width=1.16\textwidth]{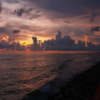}
    \end{minipage}}  & {\begin{minipage}{0.088\textwidth}
      \includegraphics[width=1.16\textwidth]{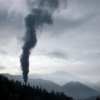}
    \end{minipage}}   & {\begin{minipage}{0.088\textwidth}
      \includegraphics[width=1.16\textwidth]{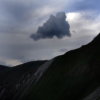}
    \end{minipage}}   &  {\begin{minipage}{0.088\textwidth}
      \includegraphics[width=1.16\textwidth]{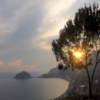}
    \end{minipage}}  & {\begin{minipage}{0.088\textwidth}
      \includegraphics[width=1.16\textwidth]{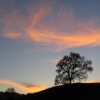}
    \end{minipage}}  & {\begin{minipage}{0.088\textwidth}
      \includegraphics[width=1.16\textwidth]{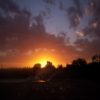}
    \end{minipage}}  & ~{\begin{minipage}{0.088\textwidth}
      \includegraphics[width=1.16\textwidth]{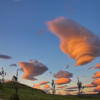}
    \end{minipage}}  & {\begin{minipage}{0.088\textwidth}
      \includegraphics[width=1.16\textwidth]{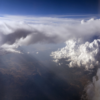}
    \end{minipage}}  & {\begin{minipage}{0.088\textwidth}
      \includegraphics[width=1.16\textwidth]{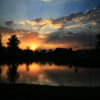}
    \end{minipage}}  \\ 
    &    &    &    &    &    &    &   &   &   & \\ \cline{2-11}
    & \multicolumn{10}{l}{HashNet}                   \\
    &  {\begin{minipage}{0.088\textwidth}
      \includegraphics[width=1.16\textwidth]{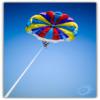}
    \end{minipage}}  &  {\begin{minipage}{0.088\textwidth}
      \includegraphics[width=1.16\textwidth]{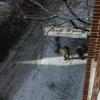}
    \end{minipage}}  & {\begin{minipage}{0.088\textwidth}
      \includegraphics[width=1.16\textwidth]{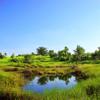}
    \end{minipage}}   & {\begin{minipage}{0.088\textwidth}
      \includegraphics[width=1.16\textwidth]{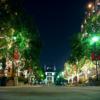}
    \end{minipage}}   &  {\begin{minipage}{0.088\textwidth}
      \includegraphics[width=1.16\textwidth]{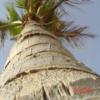}
    \end{minipage}}  & {\begin{minipage}{0.088\textwidth}
      \includegraphics[width=1.16\textwidth]{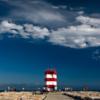}
    \end{minipage}}  & {\begin{minipage}{0.088\textwidth}
      \includegraphics[width=1.16\textwidth]{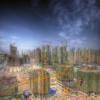}
    \end{minipage}}  & ~{\begin{minipage}{0.088\textwidth}
      \includegraphics[width=1.16\textwidth]{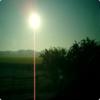}
    \end{minipage}}  & {\begin{minipage}{0.088\textwidth}
      \includegraphics[width=1.16\textwidth]{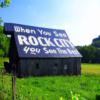}
    \end{minipage}}  & {\begin{minipage}{0.088\textwidth}
      \includegraphics[width=1.16\textwidth]{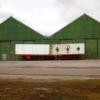}
    \end{minipage}} \\
    &    &    &    &    &    &    &   &   &   &   \\ 
    \bottomrule
    \end{tabular}%
    }

\end{strip}

  

\begin{table}
  \centering
  \caption{MAP@ALL on CIFAR-10 for image retrieval.}
  \label{tab_2}
  \begin{tabular}{p{0.14\columnwidth}p{0.14\columnwidth}<{\centering}p{0.14\columnwidth}<{\centering}p{0.14\columnwidth}<{\centering}p{0.14\columnwidth}<{\centering}}
  \toprule
  \multirow{2}{*}{Method} & \multicolumn{4}{c}{CIFAR-10 (MAP@ALL)} \\ \cline{2-5} 
  & 16 bits  & 32 bits & 48 bits & 64 bits \\ 
  \midrule
  LSH\cite{1}	&0.4443	&0.5302	&0.5839	&0.6326\\
  ITQ\cite{2}	&0.2094	&0.2355	&0.2424	&0.2535\\
  SH\cite{8}	&0.1866	&0.1900	&0.2044	&0.2020\\
  LFH\cite{43}	&0.1599	&0.1608	&0.1705	&0.1693\\
  DSDH\cite{28}	&0.7514	&0.7579	&0.7808	&0.7690\\
  HashNet\cite{12}	&0.6975	&0.7821	&0.8045&	0.8128\\
  DPSH\cite{9}	&0.7870	&0.7807	&0.7982	&0.8003\\
  DBDH\cite{44}	&0.7892	&0.7803	&0.7797	&0.7914\\
  CSQ\cite{63} &0.7761 &0.7775 &- &0.7741\\
  DSEH\cite{29}	&0.8025	&0.8130	&0.8214	&0.8301\\
  SADH	&\textbf{0.8755}	&\textbf{0.8832}	&\textbf{0.8913}	&\textbf{0.8783}\\
  \bottomrule
  \end{tabular}%
\end{table}

\begin{table}[ht]
  \centering
  \caption{MAP@5000 on NUS-WIDE for image retrieval.}
  \label{tab_3}
  \begin{tabular}{p{0.14\columnwidth}p{0.14\columnwidth}<{\centering}p{0.14\columnwidth}<{\centering}p{0.14\columnwidth}<{\centering}p{0.14\columnwidth}<{\centering}}
  \toprule
  \multirow{2}{*}{Method} & \multicolumn{4}{c}{NUS-WIDE (MAP@5000))} \\ \cline{2-5} 
  & 16 bits  & 32 bits & 48 bits & 64 bits \\ 
  \midrule
  LSH\cite{1}	&0.4443	&0.5302	&0.5839	&0.6326\\ 
  ITQ\cite{2}	&0.2094	&0.2355	&0.2424	&0.2535\\ 
  SH\cite{8}	&0.1866	&0.1900	&0.2044	&0.2020\\ 
  LFH\cite{43}	&0.1599	&0.1608	&0.1705	&0.1693\\ 
  DSDH\cite{28}	&0.7941	&0.8076	&0.8318	&0.8297\\ 
  HashNet\cite{12}	&0.7554	&0.8163	&0.8340	&0.8439\\ 
  DPSH\cite{9}	&0.8094	&0.8325	&0.8441	&\textbf{0.8520}\\ 
  DBDH\cite{44}	&0.8052	&0.8107	&0.8277	&0.8324\\ 
  CSQ\cite{63} &0.7853 &0.8213 &- &0.8316\\
  DSEH\cite{29}	&0.7319	&0.7466	&0.7602	&0.7721\\ 
  SADH	&\textbf{0.8352}	&\textbf{0.8454}	&\textbf{0.8487}	&0.8503\\ 
  \bottomrule
  \end{tabular}%
\end{table}

\begin{table}[ht]
  \centering
  \caption{MAP@ALL on MIRFLICKR-25K for image retrieval.}
  \label{tab_4}
  \begin{tabular}{p{0.14\columnwidth}p{0.14\columnwidth}<{\centering}p{0.14\columnwidth}<{\centering}p{0.14\columnwidth}<{\centering}p{0.14\columnwidth}<{\centering}}
  \toprule
  \multirow{2}{*}{Method} & \multicolumn{4}{c}{MIRFlickr-25K (MAP@ALL)} \\ \cline{2-5} 
  & 16 bits  & 32 bits & 48 bits & 64 bits \\ 
  \midrule
  DSDH\cite{28}	&0.7541	&0.7574	&0.7616	&0.7680\\ 
  HashNet\cite{12}	&0.7440	&0.7685	&0.7757	&0.7815\\ 
  DPSH\cite{9}	&0.7672	&0.7694	&0.7722	&0.7772\\ 
  DBDH\cite{44}	&0.7530	&0.7615	&0.7634	&0.7653\\ 
  CSQ\cite{63} &0.6702 &0.6735 &- &0.6843\\
  DSEH\cite{29}	&0.6832	&0.6863	&0.6974	&0.6970\\ 
  SADH	&\textbf{0.7731}	&\textbf{0.7698}	&\textbf{0.7993}	&\textbf{0.7873}\\ 
  \bottomrule
  \end{tabular}%
\end{table}
\begin{table}[ht]
  \centering
  \caption{MAP@ALL on MS-COCO for image retrieval.}
  \label{tab_5}
  \begin{tabular}{p{0.14\columnwidth}p{0.14\columnwidth}<{\centering}p{0.14\columnwidth}<{\centering}p{0.14\columnwidth}<{\centering}p{0.14\columnwidth}<{\centering}}
  \toprule
  \multirow{2}{*}{Method} & \multicolumn{4}{c}{MS-COCO (MAP@ALL)} \\ \cline{2-5} 
  & 16 bits  & 32 bits & 48 bits & 64 bits \\ 
  \midrule
  DSDH\cite{28}	&0.6093	&0.6482	&0.6615	&0.6740\\ 
  HashNet\cite{12}	&0.6873	&0.7184	&0.7301	&0.7362\\ 
  DPSH\cite{9}	&0.6610	&0.6825	&0.6887	&0.6850\\ 
  DSEH\cite{29}	&0.5897 &0.6048	&0.6133	&0.6188\\ 
  SADH	&\textbf{0.7176}	&\textbf{0.7507}	&\textbf{0.7558}	&\textbf{0.7736}\\ 
  \bottomrule
  \end{tabular}%
\end{table}


\subsubsection{Sensitivity to margin parameter}
\label{sec_432}
\begin{figure}
  \centering 
  \begin{minipage}{\columnwidth}
    \includegraphics[width=0.95\columnwidth]{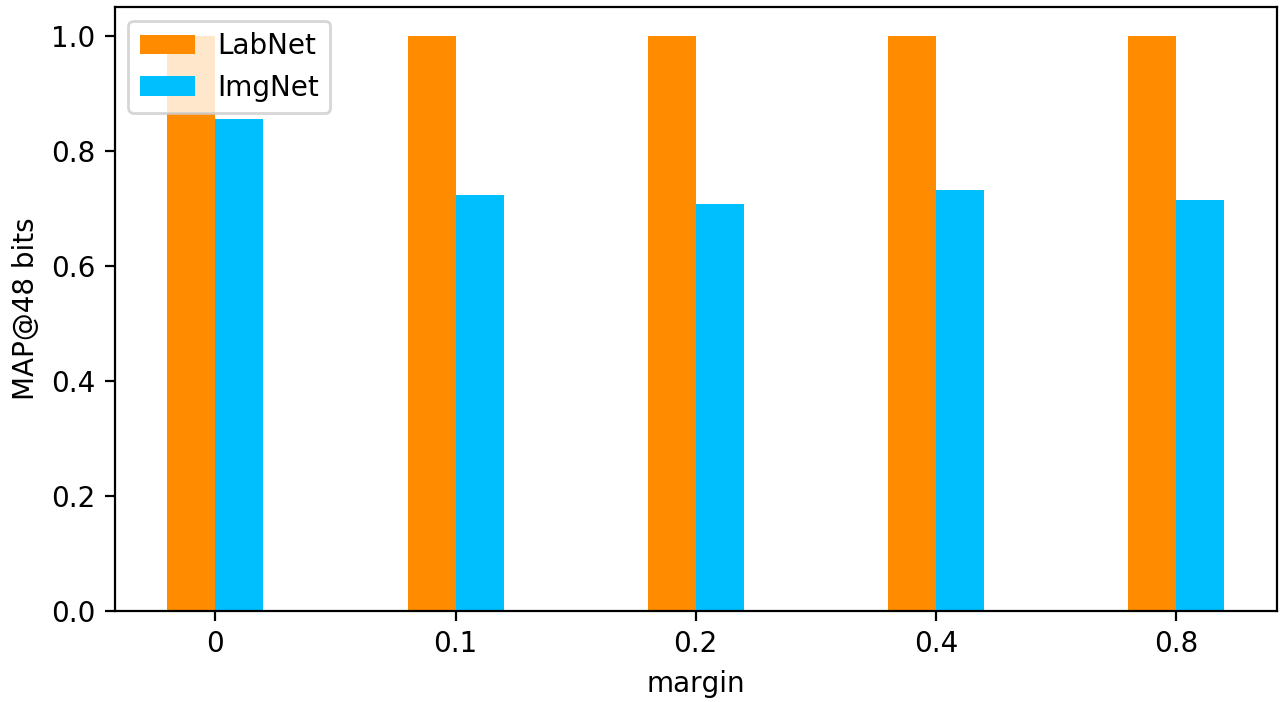}
    {\centerline{\subfigure{(a) CIFAR-10}}}
  \end{minipage}

  \begin{minipage}{0.94\columnwidth}
    \includegraphics[width=\columnwidth]{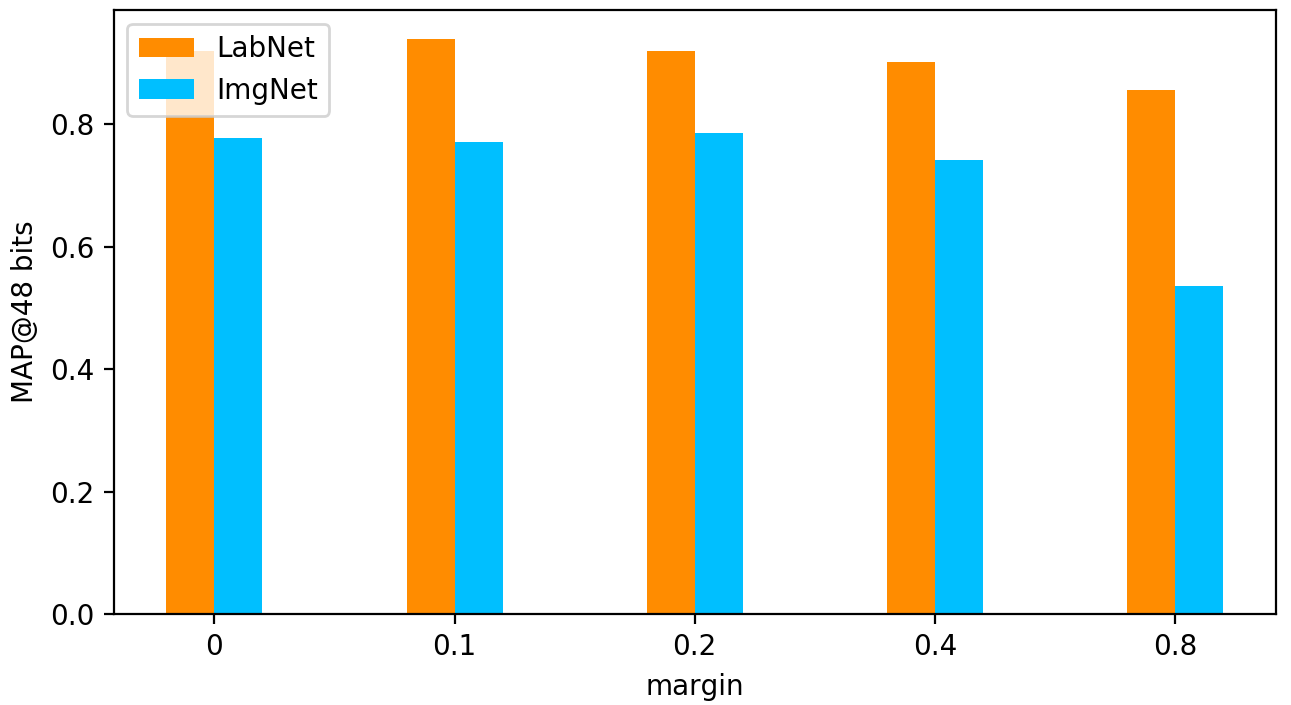}
    {\centerline{\subfigure{(b) MIRFlickr-25K}}}
  \end{minipage}
  \caption{Sensitivity analysis on the margin parameter}
  \label{fig_8}
\end{figure}

 To illustrate the earlier mentioned difference of two networks' sensitivity to margin parameter in contrastive loss, we replace the scalable margin module in Image-Network by margin constant $m$ in Semantic-Network and report their MAP with 48-bit length under different choices of $m$ on CIF- AR-10 and MIRFlicker-25K. As shown in Fig. \ref{fig_8}, we can see that under different choices of margin, Semantic-Network reveals relatively slight changes in MAP, and it’s performance is consistently robust when $m$ is relatively small, so we set $m$ as 0 for all the scenarios. While Image-Network is highly sensitive to the choice of margin with a largest MAP gap of roughly 0.14 at margin = 0 and margin = 0.2. Which to some extend reveals the significance of proper selection of margin and the feasibility of calculating margin for different item pairs rely on the hash codes generated by Semantic-Network based on the insensitivity of its performance to the selection of margin parameter.

\subsection{Empirical analysis}
Three additional experimental settings are designed and used  to further analyse SADH.

\subsubsection{Ablation study}
We investigate the impact of the different proposed modules on the retrieval performance of SADH. SADH-sym refers is built by replacing the asymmetric association between Image-Network and Semantic-Network by conventional point-to-point symmetric learning strategy, SADH-mars is built by removing the margin-scalable constraint from Image-Network, SADH-cos refers to replacing the cosine similarity module by the logarithm Maximum a Posterior (MAP) estimation of pairwise similarity loss which is used in many deep hashing approaches \cite{29,28}:
\begin{equation}\label{eq14}
  J_{s}=-\sum\limits_{i, j=1}^{m}\left(S_{i, j} H_{i}^{T} H_{j}-\log \left(1+\exp \left(H_{i}^{T} H_{j}\right)\right)\right)
\end{equation}

Results are shown on Table \ref{tab_6} for both NUS-WIDE and CIFAR-10 for hash codes of 32 bits. Considering the results, we can see that the asymmetric guidance from Semantic-Network with rich semantic information plays an essential role on the performance of our method, meanwhile the margin-scalable constraint from Image-Network itself also significantly improves retrieval accuracy. It can also be observed that when using the cosine similarity, better performance is achieved than using the MAP estimation of pairwise similarity.
\begin{table}[ht]
  \centering
  \caption{Ablation study on several modules in SADH, with MAP on NUS-WIDE
  and CIFAR-10 at hash length 32 bits}
  \label{tab_6}
  \begin{tabular}{p{0.25\columnwidth}p{0.3\columnwidth}<{\centering}p{0.3\columnwidth}<{\centering}}
  \toprule
  Methods   & \begin{tabular}[c]{@{}c@{}}NUS-WIDE\\ (MAP@5000)\end{tabular} & \begin{tabular}[c]{@{}c@{}}CIFAR-10\\ (MAP@ALL)\end{tabular} \\
  \midrule
  SADH-sym  & 0.8031                                                        & 0.8152                                                       \\
  SADH-mars & 0.8174                                                        & 0.8249                                                       \\
  SADH-cos  & 0.8168                                                        & 0.8502                                                       \\
  SADH      & \textbf{0.8454}                                               & \textbf{0.8832}                    \\ 
  \bottomrule                         
  \end{tabular}%
\end{table}

\begin{figure}[h]
  \centering 
  \begin{minipage}{\columnwidth}
    \includegraphics[width=\columnwidth]{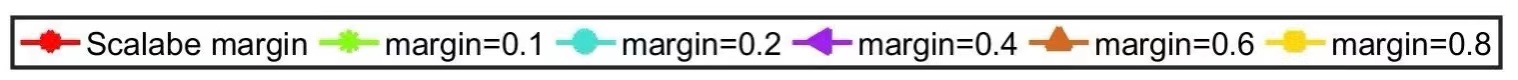}
  \end{minipage}

  \begin{minipage}{0.49\columnwidth}
    \includegraphics[width=\columnwidth]{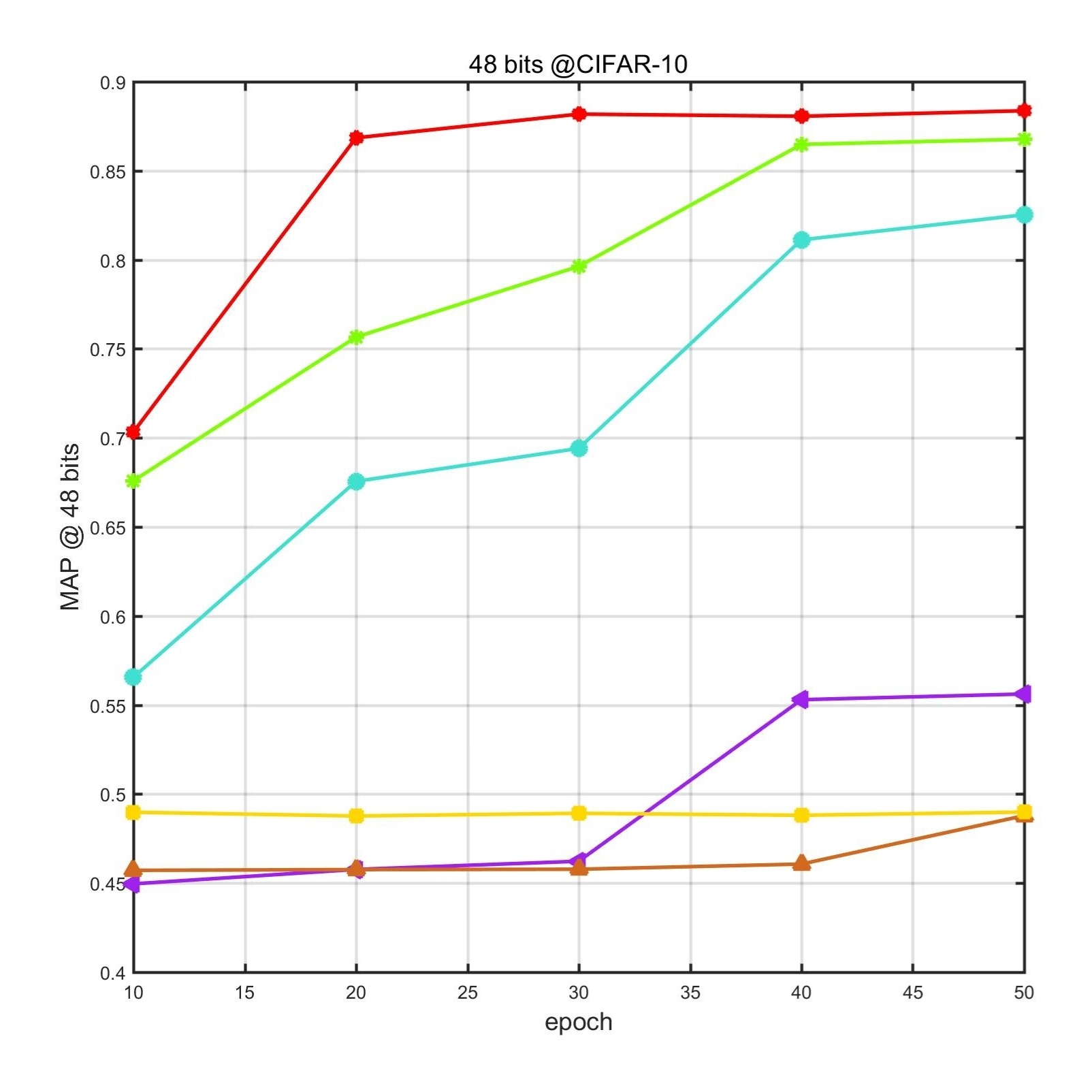}
  \end{minipage}
  \begin{minipage}{0.49\columnwidth}
    \includegraphics[width=\columnwidth]{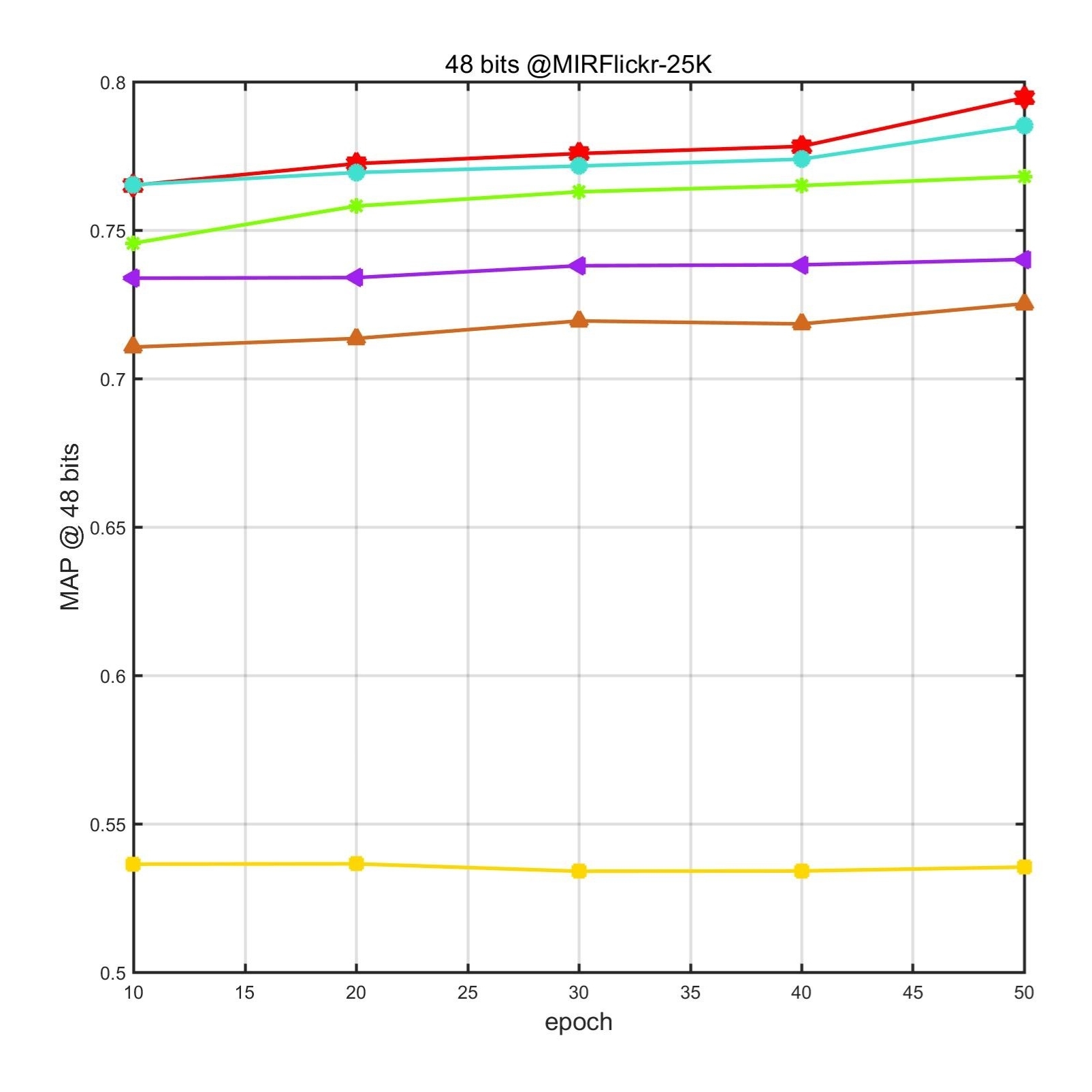}
  \end{minipage}
  \caption{Map during 50 epochs on CIFAR-10 and MIRFlickr-25K      
  with different choice of margins.}
  \label{fig_9}
\end{figure}

\vspace{10pt}
As a further demonstration of the effectiveness of the margin-scalable constraint, we compare it with several choices of single constants on our SADH. For 50 epochs, the top 5000 MAP results on MIR-Flickr25K and CIFAR-10 are given for every 10 epochs respectively. As illustrated in Fig. \ref{fig_9}. It is clear that in both the single-labeled and multi-labeled scenario, a scalable margin achieves better retrieval accuracy than using fixed margin constants. Furthermore, it is observed that on CIFAR-10, scalable margin result in faster convergence of SADH during training. 
\vspace{10pt}


\begin{figure}[h]
  \centering 
  \includegraphics[width=0.7\columnwidth]{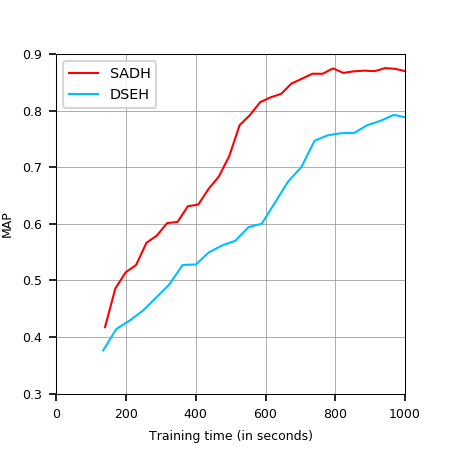}
  \caption{Training efficiency of SADH compared to DSEH on CIFAR-10.}
  \label{fig_10}
\end{figure}

\subsubsection{Training efficiency analysis}
Fig. \ref{fig_10} shows the change of MAP using 32-bit hash codes during training time of 1000 seconds, with a comparison of SADH and DSEH on CIFAR-10. We observe that SADH reduces training time by approximately two times to achieve a MAP of 0.6. Furthermore, SADH displays the tendency of convergence earlier than DSEH. SADH achieves a higher MAP than DSEH in less time. This is because Image-Network and Semantic-Network are trained jointly for multiple rounds in DSEH, with the generated hash codes and semantic features of Image-Network being supervised by same number of those generated by Semantic-Network. Whereas in SADH Semantic-Network will cease to train after one round of convergence. And the converged Semantic-Network will be utilized to produce hash code map and semantic feature map for each cases of semantic label. These maps directly supervise Image-Network with asymmetric pairwise correlation without further use of Semantic-Network.  

\subsubsection{Visualization of hash codes}
\begin{figure}[h]
  \centering 
  \begin{minipage}{0.49\columnwidth}
    \includegraphics[width=\columnwidth]{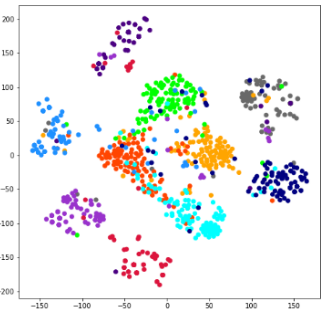}
    {\centerline{\subfigure{(a) DSDH}}}
  \end{minipage}
  \begin{minipage}{0.49\columnwidth}
    \includegraphics[width=\columnwidth]{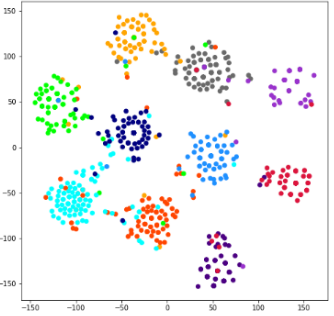}
    {\centerline{\subfigure{(b) SADH}}}
  \end{minipage}
  \caption{The t-SNE visualization of hash codes learned by\\ DSDH and SADH.}
  \label{fig_11}

\end{figure}

Fig. \ref{fig_11} is the t-SNE \cite{46} visualization of hash codes generated by DSDH and SADH on CIFAR-10, hash codes that belong to 10 different classes. Each class is assigned a different color. It can be observed that hash codes in different categories are discriminatively separated by SADH, while the hash codes generated by DSDH do not show such a clear characteristic. This is because the cosine similarity and scalable margin mechanism used in SADH can provide a more accurate inter-and-intra-class similarity preservation resulting in more discriminative hash codes in comparison to the mentioned form of pairwise similarity loss (\ref{eq14}) used in DSDH. 


\subsubsection{Heatmap visualization of focused regions}
\begin{figure*}
  \centering 
  \includegraphics[width=\textwidth]{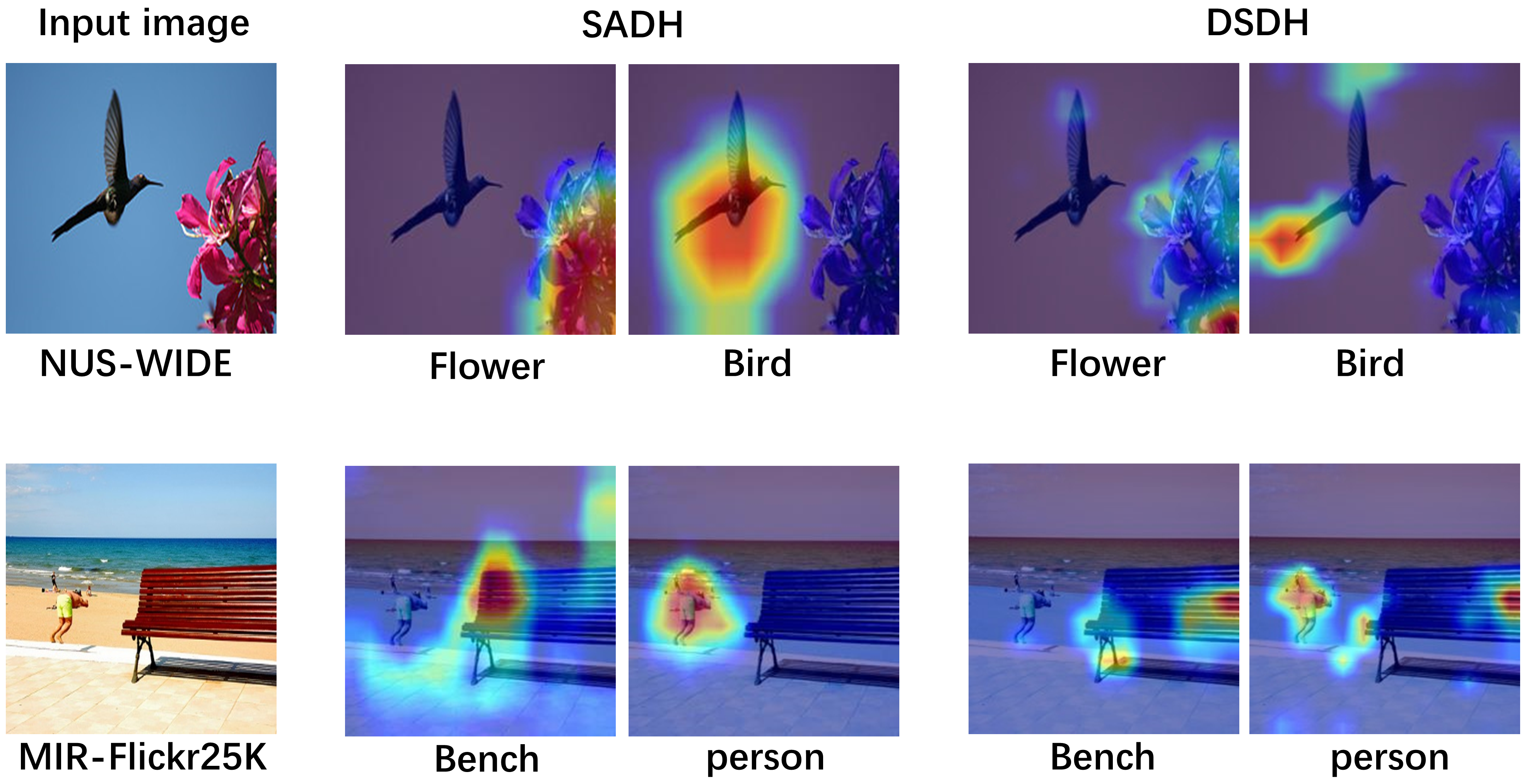}
  \caption{Grad-CAM visualization of SADH and DSDH for images sampled from multi-label benchmarks with respect to different ground-truth categories.}
  \label{fig_12}
\end{figure*}

The Grad-CAM visualization of our SADH and DSDH following\cite{64} for sampled images on NUS-WIDE and MIR-Flickr25K is illustrated in Fig. \ref{fig_12}. For each selected class of interest, Grad-CAM highlights the focused regions of convolutional feature maps. We observe that, comparing to DSDH, our SADH can correlates selected semantics with corresponding regions more accurately, which is a strong proof for robust semantic feature preserving capacity of our SADH especially for multi-label scenarios.

\begin{figure*}[h]
  \centering
\begin{minipage}{.5\textwidth}
    \centering
  \includegraphics[width=\columnwidth]{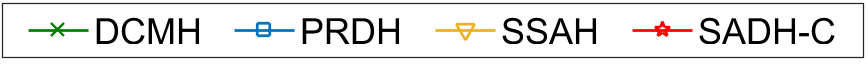}
\end{minipage}

  \includegraphics[width=\textwidth]{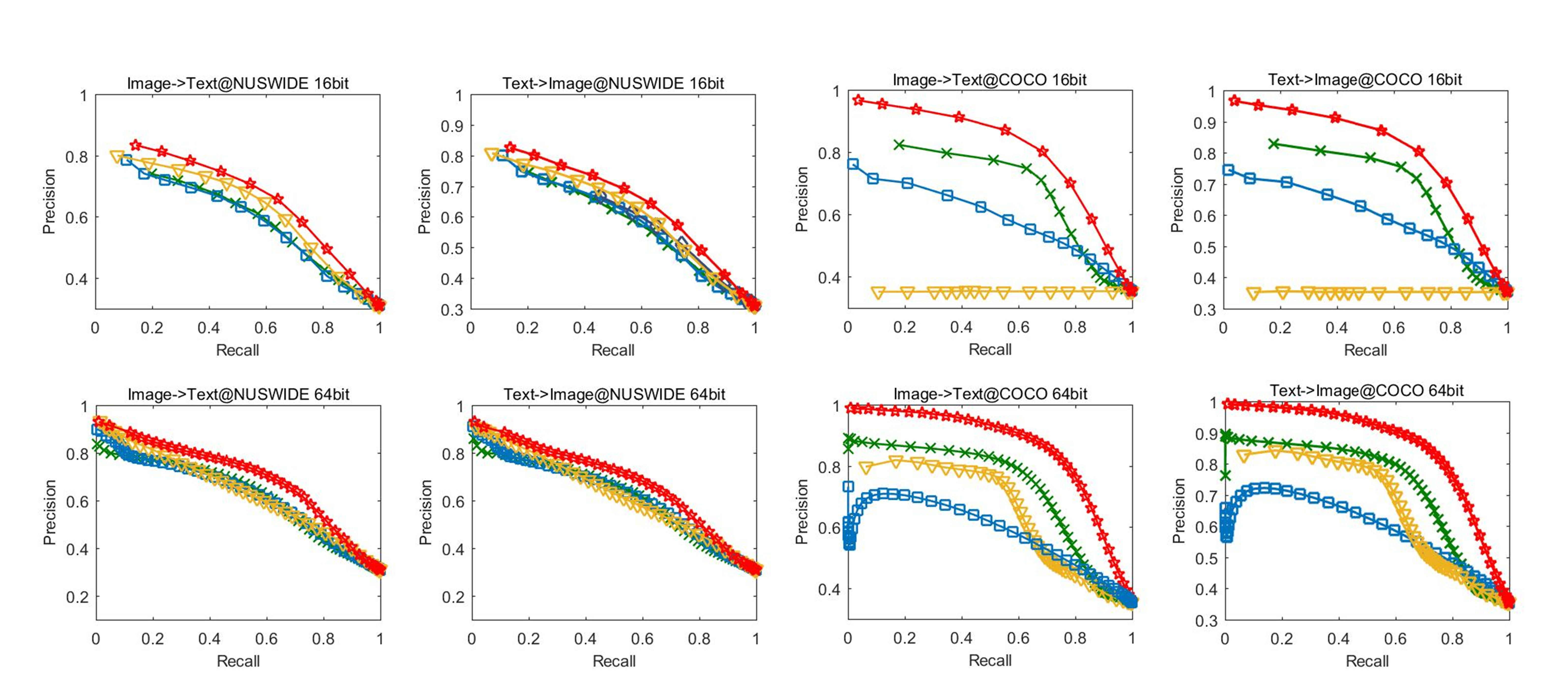}
\caption{Precision-recall curve on NUS-WIDE and\\ MS-COCO for cross-modal hashing.}
\label{fig_13}
\end{figure*}
\subsubsection{Extention: Experiments on cross-modal hashing}

As discussed earlier in \ref{sec_35}. our SADH algorithm can be seamlessly extended to cross-modal hashing. We devise a image-text cross-modal hashing framework namely SADH-c by maintaining the network architecture of Image-Network and Semantic-Network and add a 3-layer MLP network with a multi-scale fusion module to extract textual features and learn hash codes, which is the same as the TxtNet used in SSAH. Table \ref{tab_7} and Table \ref{tab_8} show the MAP result of our method and three other state-of-the-art deep supervised cross-modal hashing methods: DCMH\cite{97}, PRDH\cite{98}, SSAH\cite{62} on MS-COCO and NUS-WIDE for cross-modal retrieval between image data and text data, the according precision-recall curves are shown in Fig. \ref{fig_13}. Our approach substantially outperforms all comparison methods with particularly superior performance in MS-COCO which has 80 semantics in total, this is a strong evidence of the robustness of our method in multi-label datasets. Comparing to SSAH, which utilizes point-to-point symmetric association and logarithm Maximum a Posterior (MAP) estimation(\ref{eq14}), the remarkable performance of our proposed method is capacitated by the margin scalable pairwise constraint and asymmetric guidance mechanism.
\begin{table}
  \centering
   \caption{MAP@ALL on NUS-WIDE for cross-modal retrieval.}
  \label{tab_7}
  \begin{tabular}{p{0.14\columnwidth}p{0.14\columnwidth}<{\centering}p{0.14\columnwidth}<{\centering}p{0.14\columnwidth}<{\centering}p{0.14\columnwidth}<{\centering}}
  \hline
  \multirow{2}{*}{Task} & \multirow{2}{*}{Method} & \multicolumn{3}{c}{NUS-WIDE(MAP@ALL)} \\ \cline{3-5} 
   &  & 16bits & 48bits & 64bits \\ \hline
  \multirow{4}{*}{\begin{tabular}[c]{@{}c@{}}Image\\ to\\ Text\end{tabular}} 
   & SSAH\cite{62} & 0.6163 & 0.6278 & 0.6140 \\
   & PRDH\cite{98} & 0.5919 & 0.6059 & 0.6116 \\
   & DCMH\cite{97} & 0.5445 & 0.5597 & 0.5803 \\ 
   & SADH-c & \textbf{0.6536} & \textbf{0.6614} & \textbf{0.6663} \\ \hline
  \multirow{4}{*}{\begin{tabular}[c]{@{}c@{}}Text\\ to\\ Image\end{tabular}} 
   & SSAH\cite{62} & 0.6204 & 0.6251 & 0.6349 \\
   & PRDH\cite{98} & 0.6155 & 0.6286 & 0.6349 \\
   & DCMH\cite{97} & 0.5793 & 0.5922 & 0.6014 \\ 
   & SADH-c & \textbf{0.6748} & \textbf{0.6821} & \textbf{0.6857} \\ \hline\quad
  \end{tabular}
  
  \caption{MAP@ALL on MS-COCO for cross-modal retrieval.}
  \label{tab_8}
  \begin{tabular}{p{0.14\columnwidth}p{0.14\columnwidth}<{\centering}p{0.14\columnwidth}<{\centering}p{0.14\columnwidth}<{\centering}p{0.14\columnwidth}<{\centering}}
  \hline
  \multirow{2}{*}{Task} & \multicolumn{1}{c}{\multirow{2}{*}{Method}} & \multicolumn{3}{c}{MS-COCO(MAP@ALL)} \\ \cline{3-5} 
   & \multicolumn{1}{c}{} & 16bits & 48bits & 64bits \\ \hline
  \multirow{4}{*}{\begin{tabular}[c]{@{}c@{}}Image\\ to\\ Text\end{tabular}} 
   & SSAH\cite{62} & 0.5204 & 0.5187 & 0.5272 \\
   & PRDH\cite{98} & 0.5538 & 0.5672 & 0.5572 \\
   & DCMH\cite{97} & 0.5228 & 0.5438 & 0.5419 \\
   & SADH-c & \textbf{0.6362} & \textbf{0.6679} & \textbf{0.6929} \\ \hline
  \multirow{4}{*}{\begin{tabular}[c]{@{}c@{}}Text\\ to\\ Image\end{tabular}} 
   & SSAH\cite{62} & 0.4789 & 0.4753 & 0.4888 \\
   & PRDH\cite{98} & 0.5122 & 0.5190 & 0.5404 \\
   & DCMH\cite{97} & 0.4883 & 0.4942 & 0.5145 \\ 
   & SADH-c & \textbf{0.6347} & \textbf{0.6673} & \textbf{0.6834} \\ \hline
  \end{tabular}
  \end{table}




\section{Conclusion}
In this paper, we present a novel self-supervised asymmetric deep hashing method with a margin-scalable constraint. To improve the reliability of retrieval performance in multi-labeled scenarios, the proposed SADH preserve and refine abundant semantic information from semantic labels in two semantic dictionaries to supervise the 2nd framework Image-Network with asymmetric guidance mechanism. A margin-scalable constraint is designed to precisely search similarity information in fine-grained level. Additionally, the proposed method is seamlessly extended to cross-modal scenarios. Comprehensive empirical evidence shows that SADH outperforms several state-of-the-art methods including traditional methods and deep hashing methods on FOUR widely used benchmarks. In the future, we will explore to more detailedly investigate the proposed SADH method in deep hashing for multi-modal data retrieval.

\section*{Acknowledgements}
This work was supported by the \href{National Natural Science Foundation of China (61806168)}{National Natural Science Foundation of China (61806168)}, Fundamental Resear-ch Funds for the Central Universities (SWU117059), and Venture \& Innovation Support Program for Chongqing Overseas Returnees (CX2018075).

\bibliographystyle{model1-num-names}

\bibliography{cas-refs}

\begin{thebibliography}{89}
\expandafter\ifx\csname natexlab\endcsname\relax\def\natexlab#1{#1}\fi
\providecommand{\url}[1]{\texttt{#1}}
\providecommand{\href}[2]{#2}
\providecommand{\path}[1]{#1}
\providecommand{\DOIprefix}{doi:}
\providecommand{\ArXivprefix}{arXiv:}
\providecommand{\URLprefix}{URL: }
\providecommand{\Pubmedprefix}{pmid:}
\providecommand{\doi}[1]{\href{http://dx.doi.org/#1}{\path{#1}}}
\providecommand{\Pubmed}[1]{\href{pmid:#1}{\path{#1}}}
\providecommand{\bibinfo}[2]{#2}
\ifx\xfnm\relax \def\xfnm[#1]{\unskip,\space#1}\fi
\bibitem[{Gionis et~al.(1999)Gionis, Indyk, Motwani et~al.}]{1}
\bibinfo{author}{A.~Gionis}, \bibinfo{author}{P.~Indyk},
  \bibinfo{author}{R.~Motwani}, et~al.,
\newblock \bibinfo{title}{Similarity search in high dimensions via hashing},
\newblock in: \bibinfo{booktitle}{Vldb}, volume~\bibinfo{volume}{99},
  \bibinfo{year}{1999}, pp. \bibinfo{pages}{518--529}.
\bibitem[{Gong et~al.(2012)Gong, Lazebnik, Gordo, and Perronnin}]{2}
\bibinfo{author}{Y.~Gong}, \bibinfo{author}{S.~Lazebnik},
  \bibinfo{author}{A.~Gordo}, \bibinfo{author}{F.~Perronnin},
\newblock \bibinfo{title}{Iterative quantization: A procrustean approach to
  learning binary codes for large-scale image retrieval},
\newblock \bibinfo{journal}{IEEE transactions on pattern analysis and machine
  intelligence} \bibinfo{volume}{35} (\bibinfo{year}{2012})
  \bibinfo{pages}{2916--2929}.
\bibitem[{Salakhutdinov and Hinton(2009)}]{100}
\bibinfo{author}{R.~Salakhutdinov}, \bibinfo{author}{G.~Hinton},
\newblock \bibinfo{title}{Semantic hashing},
\newblock \bibinfo{journal}{International Journal of Approximate Reasoning}
  \bibinfo{volume}{50} (\bibinfo{year}{2009}) \bibinfo{pages}{969--978}.
\bibitem[{Wang et~al.(2014)Wang, Shen, Song, and Ji}]{31}
\bibinfo{author}{J.~Wang}, \bibinfo{author}{H.~T. Shen},
  \bibinfo{author}{J.~Song}, \bibinfo{author}{J.~Ji},
\newblock \bibinfo{title}{Hashing for similarity search: A survey},
\newblock \bibinfo{journal}{CoRR abs/1408.2927}  (\bibinfo{year}{2014}).
\bibitem[{He et~al.(2013)He, Wen, and Sun}]{5}
\bibinfo{author}{K.~He}, \bibinfo{author}{F.~Wen}, \bibinfo{author}{J.~Sun},
\newblock \bibinfo{title}{K-means hashing: An affinity-preserving quantization
  method for learning binary compact codes},
\newblock in: \bibinfo{booktitle}{Proceedings of the IEEE conference on
  computer vision and pattern recognition}, \bibinfo{year}{2013}, pp.
  \bibinfo{pages}{2938--2945}.
\bibitem[{Liu et~al.(2012)Liu, Wang, Ji, Jiang, and Chang}]{6}
\bibinfo{author}{W.~Liu}, \bibinfo{author}{J.~Wang}, \bibinfo{author}{R.~Ji},
  \bibinfo{author}{Y.-G. Jiang}, \bibinfo{author}{S.-F. Chang},
\newblock \bibinfo{title}{Supervised hashing with kernels},
\newblock in: \bibinfo{booktitle}{2012 IEEE Conference on Computer Vision and
  Pattern Recognition}, \bibinfo{organization}{IEEE}, \bibinfo{year}{2012}, pp.
  \bibinfo{pages}{2074--2081}.
\bibitem[{Kulis and Darrell(2009)}]{7}
\bibinfo{author}{B.~Kulis}, \bibinfo{author}{T.~Darrell},
\newblock \bibinfo{title}{Learning to hash with binary reconstructive
  embeddings},
\newblock in: \bibinfo{booktitle}{Advances in neural information processing
  systems}, \bibinfo{year}{2009}, pp. \bibinfo{pages}{1042--1050}.
\bibitem[{Weiss et~al.(2008)Weiss, Torralba, and Fergus}]{8}
\bibinfo{author}{Y.~Weiss}, \bibinfo{author}{A.~Torralba},
  \bibinfo{author}{R.~Fergus},
\newblock \bibinfo{title}{Spectral hashing},
\newblock \bibinfo{journal}{Advances in neural information processing systems}
  \bibinfo{volume}{21} (\bibinfo{year}{2008}) \bibinfo{pages}{1753--1760}.
\bibitem[{Norouzi and Fleet(2011)}]{59}
\bibinfo{author}{M.~Norouzi}, \bibinfo{author}{D.~J. Fleet},
\newblock \bibinfo{title}{Minimal loss hashing for compact binary codes},
\newblock in: \bibinfo{booktitle}{ICML}, \bibinfo{year}{2011}.
\bibitem[{Norouzi et~al.(2012)Norouzi, Fleet, and Salakhutdinov}]{60}
\bibinfo{author}{M.~Norouzi}, \bibinfo{author}{D.~J. Fleet},
  \bibinfo{author}{R.~R. Salakhutdinov},
\newblock \bibinfo{title}{Hamming distance metric learning},
\newblock in: \bibinfo{booktitle}{Advances in neural information processing
  systems}, \bibinfo{year}{2012}, pp. \bibinfo{pages}{1061--1069}.
\bibitem[{Li et~al.(2016)Li, Wang, and Kang}]{9}
\bibinfo{author}{W.-J. Li}, \bibinfo{author}{S.~Wang}, \bibinfo{author}{W.-C.
  Kang},
\newblock \bibinfo{title}{Feature learning based deep supervised hashing with
  pairwise labels},
\newblock \bibinfo{journal}{IJCAI}  (\bibinfo{year}{2016})
  \bibinfo{pages}{1711--1717}.
\bibitem[{Lai et~al.(2015)Lai, Pan, Liu, and Yan}]{25}
\bibinfo{author}{H.~Lai}, \bibinfo{author}{Y.~Pan}, \bibinfo{author}{Y.~Liu},
  \bibinfo{author}{S.~Yan},
\newblock \bibinfo{title}{Simultaneous feature learning and hash coding with
  deep neural networks},
\newblock in: \bibinfo{booktitle}{Proceedings of the IEEE conference on
  computer vision and pattern recognition}, \bibinfo{year}{2015}, pp.
  \bibinfo{pages}{3270--3278}.
\bibitem[{Zhu et~al.(2016)Zhu, Long, Wang, and Cao}]{11}
\bibinfo{author}{H.~Zhu}, \bibinfo{author}{M.~Long}, \bibinfo{author}{J.~Wang},
  \bibinfo{author}{Y.~Cao},
\newblock \bibinfo{title}{Deep hashing network for efficient similarity
  retrieval},
\newblock in: \bibinfo{booktitle}{Thirtieth AAAI Conference on Artificial
  Intelligence}, \bibinfo{year}{2016}.
\bibitem[{Cao et~al.(2017)Cao, Long, Wang, and Yu}]{12}
\bibinfo{author}{Z.~Cao}, \bibinfo{author}{M.~Long}, \bibinfo{author}{J.~Wang},
  \bibinfo{author}{P.~S. Yu},
\newblock \bibinfo{title}{Hashnet: Deep learning to hash by continuation},
\newblock in: \bibinfo{booktitle}{Proceedings of the IEEE international
  conference on computer vision}, \bibinfo{year}{2017}, pp.
  \bibinfo{pages}{5608--5617}.
\bibitem[{Cao et~al.(2018)Cao, Long, Liu, and Wang}]{13}
\bibinfo{author}{Y.~Cao}, \bibinfo{author}{M.~Long}, \bibinfo{author}{B.~Liu},
  \bibinfo{author}{J.~Wang},
\newblock \bibinfo{title}{Deep cauchy hashing for hamming space retrieval},
\newblock in: \bibinfo{booktitle}{Proceedings of the IEEE Conference on
  Computer Vision and Pattern Recognition}, \bibinfo{year}{2018}, pp.
  \bibinfo{pages}{1229--1237}.
\bibitem[{Liu et~al.(2018)Liu, Cao, Long, Wang, and Wang}]{14}
\bibinfo{author}{B.~Liu}, \bibinfo{author}{Y.~Cao}, \bibinfo{author}{M.~Long},
  \bibinfo{author}{J.~Wang}, \bibinfo{author}{J.~Wang},
\newblock \bibinfo{title}{Deep triplet quantization},
\newblock in: \bibinfo{booktitle}{Proceedings of the 26th ACM international
  conference on Multimedia}, \bibinfo{year}{2018}, pp.
  \bibinfo{pages}{755--763}.
\bibitem[{Chua et~al.(2009)Chua, Tang, Hong, Li, Luo, and Zheng}]{15}
\bibinfo{author}{T.-S. Chua}, \bibinfo{author}{J.~Tang},
  \bibinfo{author}{R.~Hong}, \bibinfo{author}{H.~Li}, \bibinfo{author}{Z.~Luo},
  \bibinfo{author}{Y.~Zheng},
\newblock \bibinfo{title}{Nus-wide: a real-world web image database from
  national university of singapore},
\newblock in: \bibinfo{booktitle}{Proceedings of the ACM international
  conference on image and video retrieval}, \bibinfo{year}{2009}, pp.
  \bibinfo{pages}{1--9}.
\bibitem[{Lin et~al.(2014)Lin, Maire, Belongie, Hays, Perona, Ramanan,
  Doll{\'a}r, and Zitnick}]{68}
\bibinfo{author}{T.-Y. Lin}, \bibinfo{author}{M.~Maire},
  \bibinfo{author}{S.~Belongie}, \bibinfo{author}{J.~Hays},
  \bibinfo{author}{P.~Perona}, \bibinfo{author}{D.~Ramanan},
  \bibinfo{author}{P.~Doll{\'a}r}, \bibinfo{author}{C.~L. Zitnick},
\newblock \bibinfo{title}{Microsoft coco: Common objects in context},
\newblock in: \bibinfo{editor}{D.~Fleet}, \bibinfo{editor}{T.~Pajdla},
  \bibinfo{editor}{B.~Schiele}, \bibinfo{editor}{T.~Tuytelaars} (Eds.),
  \bibinfo{booktitle}{Computer Vision -- ECCV 2014},
  \bibinfo{publisher}{Springer International Publishing},
  \bibinfo{address}{Cham}, \bibinfo{year}{2014}, pp. \bibinfo{pages}{740--755}.
\bibitem[{Huiskes and Lew(2008)}]{16}
\bibinfo{author}{M.~J. Huiskes}, \bibinfo{author}{M.~S. Lew},
\newblock \bibinfo{title}{The mir flickr retrieval evaluation},
\newblock in: \bibinfo{booktitle}{Proceedings of the 1st ACM international
  conference on Multimedia information retrieval}, \bibinfo{year}{2008}, pp.
  \bibinfo{pages}{39--43}.
\bibitem[{Jiang et~al.(2018)Jiang, Cui, and Li}]{27}
\bibinfo{author}{Q.-Y. Jiang}, \bibinfo{author}{X.~Cui}, \bibinfo{author}{W.-J.
  Li},
\newblock \bibinfo{title}{Deep discrete supervised hashing},
\newblock \bibinfo{journal}{IEEE Transactions on Image Processing}
  \bibinfo{volume}{27} (\bibinfo{year}{2018}) \bibinfo{pages}{5996--6009}.
\bibitem[{Liu et~al.(2016)Liu, Wang, Shan, and Chen}]{47}
\bibinfo{author}{H.~Liu}, \bibinfo{author}{R.~Wang}, \bibinfo{author}{S.~Shan},
  \bibinfo{author}{X.~Chen},
\newblock \bibinfo{title}{Deep supervised hashing for fast image retrieval},
\newblock in: \bibinfo{booktitle}{Proceedings of the IEEE conference on
  computer vision and pattern recognition}, \bibinfo{year}{2016}, pp.
  \bibinfo{pages}{2064--2072}.
\bibitem[{Jun et~al.(2019)Jun, Ko, Kim, Kim, and Kim}]{67}
\bibinfo{author}{H.~Jun}, \bibinfo{author}{B.~Ko}, \bibinfo{author}{Y.~Kim},
  \bibinfo{author}{I.~Kim}, \bibinfo{author}{J.~Kim},
\newblock \bibinfo{title}{Combination of multiple global descriptors for image
  retrieval},
\newblock \bibinfo{journal}{arXiv preprint arXiv:1903.10663}
  (\bibinfo{year}{2019}).
\bibitem[{Xia et~al.(2014)Xia, Pan, Lai, Liu, and Yan}]{42}
\bibinfo{author}{R.~Xia}, \bibinfo{author}{Y.~Pan}, \bibinfo{author}{H.~Lai},
  \bibinfo{author}{C.~Liu}, \bibinfo{author}{S.~Yan},
\newblock \bibinfo{title}{Supervised hashing for image retrieval via image
  representation learning.},
\newblock in: \bibinfo{booktitle}{AAAI}, volume~\bibinfo{volume}{1},
  \bibinfo{year}{2014}, p.~\bibinfo{pages}{2}.
\bibitem[{Li et~al.(2017)Li, Sun, He, and Tan}]{28}
\bibinfo{author}{Q.~Li}, \bibinfo{author}{Z.~Sun}, \bibinfo{author}{R.~He},
  \bibinfo{author}{T.~Tan},
\newblock \bibinfo{title}{Deep supervised discrete hashing},
\newblock in: \bibinfo{editor}{I.~Guyon}, \bibinfo{editor}{U.~V. Luxburg},
  \bibinfo{editor}{S.~Bengio}, \bibinfo{editor}{H.~Wallach},
  \bibinfo{editor}{R.~Fergus}, \bibinfo{editor}{S.~Vishwanathan},
  \bibinfo{editor}{R.~Garnett} (Eds.), \bibinfo{booktitle}{Advances in Neural
  Information Processing Systems}, volume~\bibinfo{volume}{30},
  \bibinfo{publisher}{Curran Associates, Inc.}, \bibinfo{year}{2017}.
  \URLprefix
  \url{https://proceedings.neurips.cc/paper/2017/file/e94f63f579e05cb49c05c2d050ead9c0-Paper.pdf}.
\bibitem[{Shen et~al.(2015)Shen, Shen, Liu, and Tao~Shen}]{65}
\bibinfo{author}{F.~Shen}, \bibinfo{author}{C.~Shen}, \bibinfo{author}{W.~Liu},
  \bibinfo{author}{H.~Tao~Shen},
\newblock \bibinfo{title}{Supervised discrete hashing},
\newblock in: \bibinfo{booktitle}{Proceedings of the IEEE conference on
  computer vision and pattern recognition}, \bibinfo{year}{2015}, pp.
  \bibinfo{pages}{37--45}.
\bibitem[{Da et~al.(2018)Da, Meng, Xiang, Ding, Xu, Yang, and Pan}]{66}
\bibinfo{author}{C.~Da}, \bibinfo{author}{G.~Meng}, \bibinfo{author}{S.~Xiang},
  \bibinfo{author}{K.~Ding}, \bibinfo{author}{S.~Xu},
  \bibinfo{author}{Q.~Yang}, \bibinfo{author}{C.~Pan},
\newblock \bibinfo{title}{Nonlinear asymmetric multi-valued hashing},
\newblock \bibinfo{journal}{IEEE transactions on pattern analysis and machine
  intelligence} \bibinfo{volume}{41} (\bibinfo{year}{2018})
  \bibinfo{pages}{2660--2676}.
\bibitem[{Shrivastava and Li(2014)}]{88}
\bibinfo{author}{A.~Shrivastava}, \bibinfo{author}{P.~Li},
\newblock \bibinfo{title}{Asymmetric lsh (alsh) for sublinear time maximum
  inner product search (mips)},
\newblock \bibinfo{journal}{arXiv preprint arXiv:1405.5869}
  (\bibinfo{year}{2014}).
\bibitem[{Zhang et~al.(2019)Zhang, Lai, Huang, Wong, Xie, Liu, and Shao}]{89}
\bibinfo{author}{Z.~Zhang}, \bibinfo{author}{Z.~Lai},
  \bibinfo{author}{Z.~Huang}, \bibinfo{author}{W.~K. Wong},
  \bibinfo{author}{G.-S. Xie}, \bibinfo{author}{L.~Liu},
  \bibinfo{author}{L.~Shao},
\newblock \bibinfo{title}{Scalable supervised asymmetric hashing with semantic
  and latent factor embedding},
\newblock \bibinfo{journal}{IEEE Transactions on Image Processing}
  \bibinfo{volume}{28} (\bibinfo{year}{2019}) \bibinfo{pages}{4803--4818}.
\bibitem[{Neyshabur et~al.(2013)Neyshabur, Yadollahpour, Makarychev,
  Salakhutdinov, and Srebro}]{90}
\bibinfo{author}{B.~Neyshabur}, \bibinfo{author}{P.~Yadollahpour},
  \bibinfo{author}{Y.~Makarychev}, \bibinfo{author}{R.~Salakhutdinov},
  \bibinfo{author}{N.~Srebro},
\newblock \bibinfo{title}{The power of asymmetry in binary hashing},
\newblock \bibinfo{journal}{arXiv preprint arXiv:1311.7662}
  (\bibinfo{year}{2013}).
\bibitem[{Jiang and Li(2018)}]{32}
\bibinfo{author}{Q.-Y. Jiang}, \bibinfo{author}{W.-J. Li},
\newblock \bibinfo{title}{Asymmetric deep supervised hashing},
\newblock \bibinfo{journal}{AAAI}  (\bibinfo{year}{2018})
  \bibinfo{pages}{3342--3349}.
\bibitem[{Indyk and Motwani(1998)}]{69}
\bibinfo{author}{P.~Indyk}, \bibinfo{author}{R.~Motwani},
\newblock \bibinfo{title}{Approximate nearest neighbors: towards removing the
  curse of dimensionality},
\newblock in: \bibinfo{booktitle}{Proceedings of the thirtieth annual ACM
  symposium on Theory of computing}, \bibinfo{year}{1998}, pp.
  \bibinfo{pages}{604--613}.
\bibitem[{Charikar(2002)}]{70}
\bibinfo{author}{M.~S. Charikar},
\newblock \bibinfo{title}{Similarity estimation techniques from rounding
  algorithms},
\newblock in: \bibinfo{booktitle}{Proceedings of the thiry-fourth annual ACM
  symposium on Theory of computing}, \bibinfo{year}{2002}, pp.
  \bibinfo{pages}{380--388}.
\bibitem[{Wang et~al.(2018)Wang, Liu, Xia, Ramamohanarao, and Tao}]{71}
\bibinfo{author}{B.~Wang}, \bibinfo{author}{X.~Liu}, \bibinfo{author}{K.~Xia},
  \bibinfo{author}{K.~Ramamohanarao}, \bibinfo{author}{D.~Tao},
\newblock \bibinfo{title}{Random angular projection for fast nearest subspace
  search},
\newblock in: \bibinfo{booktitle}{Pacific Rim Conference on Multimedia},
  \bibinfo{organization}{Springer}, \bibinfo{year}{2018}, pp.
  \bibinfo{pages}{15--26}.
\bibitem[{Ji et~al.(2015)Ji, Li, Tian, Yan, and Zhang}]{72}
\bibinfo{author}{J.~Ji}, \bibinfo{author}{J.~Li}, \bibinfo{author}{Q.~Tian},
  \bibinfo{author}{S.~Yan}, \bibinfo{author}{B.~Zhang},
\newblock \bibinfo{title}{Angular-similarity-preserving binary signatures for
  linear subspaces},
\newblock \bibinfo{journal}{IEEE Transactions on Image Processing}
  \bibinfo{volume}{24} (\bibinfo{year}{2015}) \bibinfo{pages}{4372--4380}.
\bibitem[{Xu et~al.(2021)Xu, Liu, Wang, Tao, Xia, and Cao}]{73}
\bibinfo{author}{Y.~Xu}, \bibinfo{author}{X.~Liu}, \bibinfo{author}{B.~Wang},
  \bibinfo{author}{R.~Tao}, \bibinfo{author}{K.~Xia}, \bibinfo{author}{X.~Cao},
\newblock \bibinfo{title}{Fast nearest subspace search via random angular
  hashing},
\newblock \bibinfo{journal}{IEEE Transactions on Multimedia}
  \bibinfo{volume}{23} (\bibinfo{year}{2021}) \bibinfo{pages}{342--352}.
\bibitem[{Gong et~al.(2012)Gong, Lazebnik, Gordo, and Perronnin}]{4}
\bibinfo{author}{Y.~Gong}, \bibinfo{author}{S.~Lazebnik},
  \bibinfo{author}{A.~Gordo}, \bibinfo{author}{F.~Perronnin},
\newblock \bibinfo{title}{Iterative quantization: A procrustean approach to
  learning binary codes for large-scale image retrieval},
\newblock \bibinfo{journal}{IEEE transactions on pattern analysis and machine
  intelligence} \bibinfo{volume}{35} (\bibinfo{year}{2012})
  \bibinfo{pages}{2916--2929}.
\bibitem[{Lu et~al.(2016)Lu, Zheng, and Li}]{74}
\bibinfo{author}{X.~Lu}, \bibinfo{author}{X.~Zheng}, \bibinfo{author}{X.~Li},
\newblock \bibinfo{title}{Latent semantic minimal hashing for image retrieval},
\newblock \bibinfo{journal}{IEEE Transactions on Image Processing}
  \bibinfo{volume}{26} (\bibinfo{year}{2016}) \bibinfo{pages}{355--368}.
\bibitem[{Yuan et~al.(2017)Yuan, Deng, and Hu}]{19}
\bibinfo{author}{T.~Yuan}, \bibinfo{author}{W.~Deng}, \bibinfo{author}{J.~Hu},
\newblock \bibinfo{title}{Distortion minimization hashing},
\newblock \bibinfo{journal}{IEEE Access} \bibinfo{volume}{5}
  (\bibinfo{year}{2017}) \bibinfo{pages}{23425--23435}.
\bibitem[{He et~al.(2016)He, Zhang, Ren, and Sun}]{21}
\bibinfo{author}{K.~He}, \bibinfo{author}{X.~Zhang}, \bibinfo{author}{S.~Ren},
  \bibinfo{author}{J.~Sun},
\newblock \bibinfo{title}{Deep residual learning for image recognition},
\newblock in: \bibinfo{booktitle}{Proceedings of the IEEE conference on
  computer vision and pattern recognition}, \bibinfo{year}{2016}, pp.
  \bibinfo{pages}{770--778}.
\bibitem[{Joly and Buisson(2011)}]{85}
\bibinfo{author}{A.~Joly}, \bibinfo{author}{O.~Buisson},
\newblock \bibinfo{title}{Random maximum margin hashing},
\newblock in: \bibinfo{booktitle}{CVPR 2011}, \bibinfo{organization}{IEEE},
  \bibinfo{year}{2011}, pp. \bibinfo{pages}{873--880}.
\bibitem[{Yang et~al.(2016)Yang, Bai, Liu, Wang, Bai, Zhou, and Tang}]{86}
\bibinfo{author}{H.~Yang}, \bibinfo{author}{X.~Bai}, \bibinfo{author}{Y.~Liu},
  \bibinfo{author}{Y.~Wang}, \bibinfo{author}{L.~Bai},
  \bibinfo{author}{J.~Zhou}, \bibinfo{author}{W.~Tang},
\newblock \bibinfo{title}{Maximum margin hashing with supervised information},
\newblock \bibinfo{journal}{Multimedia Tools and Applications}
  \bibinfo{volume}{75} (\bibinfo{year}{2016}) \bibinfo{pages}{3955--3971}.
\bibitem[{Krizhevsky et~al.(2017)Krizhevsky, Sutskever, and Hinton}]{23}
\bibinfo{author}{A.~Krizhevsky}, \bibinfo{author}{I.~Sutskever},
  \bibinfo{author}{G.~E. Hinton},
\newblock \bibinfo{title}{Imagenet classification with deep convolutional
  neural networks},
\newblock \bibinfo{journal}{Communications of the ACM} \bibinfo{volume}{60}
  (\bibinfo{year}{2017}) \bibinfo{pages}{84--90}.
\bibitem[{Tang et~al.(2018)Tang, Li, and Zhu}]{24}
\bibinfo{author}{J.~Tang}, \bibinfo{author}{Z.~Li}, \bibinfo{author}{X.~Zhu},
\newblock \bibinfo{title}{Supervised deep hashing for scalable face image
  retrieval},
\newblock \bibinfo{journal}{Pattern Recognition} \bibinfo{volume}{75}
  (\bibinfo{year}{2018}) \bibinfo{pages}{25--32}.
\bibitem[{Guo et~al.(2016)Guo, Liu, Oerlemans, Lao, Wu, and Lew}]{53}
\bibinfo{author}{Y.~Guo}, \bibinfo{author}{Y.~Liu},
  \bibinfo{author}{A.~Oerlemans}, \bibinfo{author}{S.~Lao},
  \bibinfo{author}{S.~Wu}, \bibinfo{author}{M.~S. Lew},
\newblock \bibinfo{title}{Deep learning for visual understanding: A review},
\newblock \bibinfo{journal}{Neurocomputing} \bibinfo{volume}{187}
  (\bibinfo{year}{2016}) \bibinfo{pages}{27--48}.
\bibitem[{Li et~al.(2015)Li, Lin, Shen, Brandt, and Hua}]{54}
\bibinfo{author}{H.~Li}, \bibinfo{author}{Z.~Lin}, \bibinfo{author}{X.~Shen},
  \bibinfo{author}{J.~Brandt}, \bibinfo{author}{G.~Hua},
\newblock \bibinfo{title}{A convolutional neural network cascade for face
  detection},
\newblock in: \bibinfo{booktitle}{Proceedings of the IEEE conference on
  computer vision and pattern recognition}, \bibinfo{year}{2015}, pp.
  \bibinfo{pages}{5325--5334}.
\bibitem[{Liu et~al.(2015)Liu, Li, Luo, Loy, and Tang}]{55}
\bibinfo{author}{Z.~Liu}, \bibinfo{author}{X.~Li}, \bibinfo{author}{P.~Luo},
  \bibinfo{author}{C.-C. Loy}, \bibinfo{author}{X.~Tang},
\newblock \bibinfo{title}{Semantic image segmentation via deep parsing
  network},
\newblock in: \bibinfo{booktitle}{Proceedings of the IEEE international
  conference on computer vision}, \bibinfo{year}{2015}, pp.
  \bibinfo{pages}{1377--1385}.
\bibitem[{He et~al.(2015)He, Zhang, Ren, and Sun}]{56}
\bibinfo{author}{K.~He}, \bibinfo{author}{X.~Zhang}, \bibinfo{author}{S.~Ren},
  \bibinfo{author}{J.~Sun},
\newblock \bibinfo{title}{Spatial pyramid pooling in deep convolutional
  networks for visual recognition},
\newblock \bibinfo{journal}{IEEE transactions on pattern analysis and machine
  intelligence} \bibinfo{volume}{37} (\bibinfo{year}{2015})
  \bibinfo{pages}{1904--1916}.
\bibitem[{Girshick(2015)}]{57}
\bibinfo{author}{R.~Girshick},
\newblock \bibinfo{title}{Fast r-cnn},
\newblock in: \bibinfo{booktitle}{Proceedings of the IEEE international
  conference on computer vision}, \bibinfo{year}{2015}, pp.
  \bibinfo{pages}{1440--1448}.
\bibitem[{Ren et~al.(2015)Ren, He, Girshick, and Sun}]{58}
\bibinfo{author}{S.~Ren}, \bibinfo{author}{K.~He},
  \bibinfo{author}{R.~Girshick}, \bibinfo{author}{J.~Sun},
\newblock \bibinfo{title}{Faster r-cnn: Towards real-time object detection with
  region proposal networks},
\newblock in: \bibinfo{booktitle}{Advances in neural information processing
  systems}, \bibinfo{year}{2015}, pp. \bibinfo{pages}{91--99}.
\bibitem[{Wu et~al.(2017)Wu, Oerlemans, Bakker, and Lew}]{add55}
\bibinfo{author}{S.~Wu}, \bibinfo{author}{A.~Oerlemans}, \bibinfo{author}{E.~M.
  Bakker}, \bibinfo{author}{M.~S. Lew},
\newblock \bibinfo{title}{Deep binary codes for large scale image retrieval},
\newblock \bibinfo{journal}{Neurocomputing} \bibinfo{volume}{257}
  (\bibinfo{year}{2017}) \bibinfo{pages}{5--15}. \bibinfo{note}{Machine
  Learning and Signal Processing for Big Multimedia Analysis}.
\bibitem[{Guo et~al.(2016)Guo, Liu, Oerlemans, Lao, Wu, and Lew}]{add56}
\bibinfo{author}{Y.~Guo}, \bibinfo{author}{Y.~Liu},
  \bibinfo{author}{A.~Oerlemans}, \bibinfo{author}{S.~Lao},
  \bibinfo{author}{S.~Wu}, \bibinfo{author}{M.~S. Lew},
\newblock \bibinfo{title}{Deep learning for visual understanding: A review},
\newblock \bibinfo{journal}{Neurocomputing} \bibinfo{volume}{187}
  (\bibinfo{year}{2016}) \bibinfo{pages}{27--48}. \bibinfo{note}{Recent
  Developments on Deep Big Vision}.
\bibitem[{Wang et~al.(2020)Wang, Zou, Bakker, and Wu}]{add57}
\bibinfo{author}{X.~Wang}, \bibinfo{author}{X.~Zou}, \bibinfo{author}{E.~M.
  Bakker}, \bibinfo{author}{S.~Wu},
\newblock \bibinfo{title}{Self-constraining and attention-based hashing network
  for bit-scalable cross-modal retrieval},
\newblock \bibinfo{journal}{Neurocomputing} \bibinfo{volume}{400}
  (\bibinfo{year}{2020}) \bibinfo{pages}{255--271}.
\bibitem[{Lai et~al.(2016)Lai, Yan, Shu, Wei, and Yan}]{75}
\bibinfo{author}{H.~Lai}, \bibinfo{author}{P.~Yan}, \bibinfo{author}{X.~Shu},
  \bibinfo{author}{Y.~Wei}, \bibinfo{author}{S.~Yan},
\newblock \bibinfo{title}{Instance-aware hashing for multi-label image
  retrieval},
\newblock \bibinfo{journal}{IEEE Transactions on Image Processing}
  \bibinfo{volume}{25} (\bibinfo{year}{2016}) \bibinfo{pages}{2469--2479}.
\bibitem[{Lin et~al.(2015)Lin, Yang, Hsiao, and Chen}]{76}
\bibinfo{author}{K.~Lin}, \bibinfo{author}{H.-F. Yang}, \bibinfo{author}{J.-H.
  Hsiao}, \bibinfo{author}{C.-S. Chen},
\newblock \bibinfo{title}{Deep learning of binary hash codes for fast image
  retrieval},
\newblock in: \bibinfo{booktitle}{Proceedings of the IEEE conference on
  computer vision and pattern recognition workshops}, \bibinfo{year}{2015}, pp.
  \bibinfo{pages}{27--35}.
\bibitem[{Yang et~al.(2017)Yang, Lin, and Chen}]{77}
\bibinfo{author}{H.-F. Yang}, \bibinfo{author}{K.~Lin}, \bibinfo{author}{C.-S.
  Chen},
\newblock \bibinfo{title}{Supervised learning of semantics-preserving hash via
  deep convolutional neural networks},
\newblock \bibinfo{journal}{IEEE transactions on pattern analysis and machine
  intelligence} \bibinfo{volume}{40} (\bibinfo{year}{2017})
  \bibinfo{pages}{437--451}.
\bibitem[{Yao et~al.(2016)Yao, Long, Mei, and Rui}]{78}
\bibinfo{author}{T.~Yao}, \bibinfo{author}{F.~Long}, \bibinfo{author}{T.~Mei},
  \bibinfo{author}{Y.~Rui},
\newblock \bibinfo{title}{Deep semantic-preserving and ranking-based hashing
  for image retrieval.},
\newblock in: \bibinfo{booktitle}{IJCAI}, volume~\bibinfo{volume}{1},
  \bibinfo{year}{2016}, p.~\bibinfo{pages}{4}.
\bibitem[{Shen et~al.(2015)Shen, Shen, Liu, and Tao~Shen}]{79}
\bibinfo{author}{F.~Shen}, \bibinfo{author}{C.~Shen}, \bibinfo{author}{W.~Liu},
  \bibinfo{author}{H.~Tao~Shen},
\newblock \bibinfo{title}{Supervised discrete hashing},
\newblock in: \bibinfo{booktitle}{Proceedings of the IEEE conference on
  computer vision and pattern recognition}, \bibinfo{year}{2015}, pp.
  \bibinfo{pages}{37--45}.
\bibitem[{Xie et~al.(2020)Xie, Deng, Li, Liu, and Tao}]{80}
\bibinfo{author}{D.~Xie}, \bibinfo{author}{C.~Deng}, \bibinfo{author}{C.~Li},
  \bibinfo{author}{X.~Liu}, \bibinfo{author}{D.~Tao},
\newblock \bibinfo{title}{Multi-task consistency-preserving adversarial hashing
  for cross-modal retrieval},
\newblock \bibinfo{journal}{IEEE Transactions on Image Processing}
  \bibinfo{volume}{29} (\bibinfo{year}{2020}) \bibinfo{pages}{3626--3637}.
\bibitem[{Da et~al.(2018)Da, Meng, Xiang, Ding, Xu, Yang, and Pan}]{81}
\bibinfo{author}{C.~Da}, \bibinfo{author}{G.~Meng}, \bibinfo{author}{S.~Xiang},
  \bibinfo{author}{K.~Ding}, \bibinfo{author}{S.~Xu},
  \bibinfo{author}{Q.~Yang}, \bibinfo{author}{C.~Pan},
\newblock \bibinfo{title}{Nonlinear asymmetric multi-valued hashing},
\newblock \bibinfo{journal}{IEEE transactions on pattern analysis and machine
  intelligence} \bibinfo{volume}{41} (\bibinfo{year}{2018})
  \bibinfo{pages}{2660--2676}.
\bibitem[{Li et~al.(2018)Li, Li, Deng, Liu, and Gao}]{29}
\bibinfo{author}{N.~Li}, \bibinfo{author}{C.~Li}, \bibinfo{author}{C.~Deng},
  \bibinfo{author}{X.~Liu}, \bibinfo{author}{X.~Gao},
\newblock \bibinfo{title}{Deep joint semantic-embedding hashing.},
\newblock in: \bibinfo{booktitle}{IJCAI}, \bibinfo{year}{2018}, pp.
  \bibinfo{pages}{2397--2403}.
\bibitem[{Zhao et~al.(2015)Zhao, Huang, Wang, and Tan}]{82}
\bibinfo{author}{F.~Zhao}, \bibinfo{author}{Y.~Huang},
  \bibinfo{author}{L.~Wang}, \bibinfo{author}{T.~Tan},
\newblock \bibinfo{title}{Deep semantic ranking based hashing for multi-label
  image retrieval},
\newblock in: \bibinfo{booktitle}{Proceedings of the IEEE conference on
  computer vision and pattern recognition}, \bibinfo{year}{2015}, pp.
  \bibinfo{pages}{1556--1564}.
\bibitem[{Zhang et~al.(2019)Zhang, Zou, Lin, Chen, and Wang}]{83}
\bibinfo{author}{Z.~Zhang}, \bibinfo{author}{Q.~Zou}, \bibinfo{author}{Y.~Lin},
  \bibinfo{author}{L.~Chen}, \bibinfo{author}{S.~Wang},
\newblock \bibinfo{title}{Improved deep hashing with soft pairwise similarity
  for multi-label image retrieval},
\newblock \bibinfo{journal}{IEEE Transactions on Multimedia}
  \bibinfo{volume}{22} (\bibinfo{year}{2019}) \bibinfo{pages}{540--553}.
\bibitem[{Hadsell et~al.(2006)Hadsell, Chopra, and LeCun}]{84}
\bibinfo{author}{R.~Hadsell}, \bibinfo{author}{S.~Chopra},
  \bibinfo{author}{Y.~LeCun},
\newblock \bibinfo{title}{Dimensionality reduction by learning an invariant
  mapping},
\newblock in: \bibinfo{booktitle}{2006 IEEE Computer Society Conference on
  Computer Vision and Pattern Recognition (CVPR'06)},
  volume~\bibinfo{volume}{2}, \bibinfo{organization}{IEEE},
  \bibinfo{year}{2006}, pp. \bibinfo{pages}{1735--1742}.
\bibitem[{Kang et~al.(2019)Kang, Cao, Long, Wang, and Yu}]{87}
\bibinfo{author}{R.~Kang}, \bibinfo{author}{Y.~Cao}, \bibinfo{author}{M.~Long},
  \bibinfo{author}{J.~Wang}, \bibinfo{author}{P.~S. Yu},
\newblock \bibinfo{title}{Maximum-margin hamming hashing},
\newblock in: \bibinfo{booktitle}{Proceedings of the IEEE/CVF International
  Conference on Computer Vision}, \bibinfo{year}{2019}, pp.
  \bibinfo{pages}{8252--8261}.
\bibitem[{Li et~al.(2018)Li, Deng, Li, Liu, Gao, and Tao}]{62}
\bibinfo{author}{C.~Li}, \bibinfo{author}{C.~Deng}, \bibinfo{author}{N.~Li},
  \bibinfo{author}{W.~Liu}, \bibinfo{author}{X.~Gao}, \bibinfo{author}{D.~Tao},
\newblock \bibinfo{title}{Self-supervised adversarial hashing networks for
  cross-modal retrieval},
\newblock in: \bibinfo{booktitle}{Proceedings of the IEEE conference on
  computer vision and pattern recognition}, \bibinfo{year}{2018}, pp.
  \bibinfo{pages}{4242--4251}.
\bibitem[{Gu et~al.(2019)Gu, Gu, Gu, Li, Xiong, and Wang}]{34}
\bibinfo{author}{W.~Gu}, \bibinfo{author}{X.~Gu}, \bibinfo{author}{J.~Gu},
  \bibinfo{author}{B.~Li}, \bibinfo{author}{Z.~Xiong},
  \bibinfo{author}{W.~Wang},
\newblock \bibinfo{title}{Adversary guided asymmetric hashing for cross-modal
  retrieval},
\newblock in: \bibinfo{booktitle}{Proceedings of the 2019 on International
  Conference on Multimedia Retrieval}, \bibinfo{year}{2019}, pp.
  \bibinfo{pages}{159--167}.
\bibitem[{Song et~al.(2013)Song, Yang, Yang, Huang, and Shen}]{91}
\bibinfo{author}{J.~Song}, \bibinfo{author}{Y.~Yang},
  \bibinfo{author}{Y.~Yang}, \bibinfo{author}{Z.~Huang}, \bibinfo{author}{H.~T.
  Shen},
\newblock \bibinfo{title}{Inter-media hashing for large-scale retrieval from
  heterogeneous data sources},
\newblock in: \bibinfo{booktitle}{Proceedings of the 2013 ACM SIGMOD
  International Conference on Management of Data}, \bibinfo{year}{2013}, pp.
  \bibinfo{pages}{785--796}.
\bibitem[{Zhou et~al.(2014)Zhou, Ding, and Guo}]{92}
\bibinfo{author}{J.~Zhou}, \bibinfo{author}{G.~Ding}, \bibinfo{author}{Y.~Guo},
\newblock \bibinfo{title}{Latent semantic sparse hashing for cross-modal
  similarity search},
\newblock in: \bibinfo{booktitle}{Proceedings of the 37th international ACM
  SIGIR conference on Research \& development in information retrieval},
  \bibinfo{year}{2014}, pp. \bibinfo{pages}{415--424}.
\bibitem[{Ding et~al.(2016)Ding, Guo, Zhou, and Gao}]{93}
\bibinfo{author}{G.~Ding}, \bibinfo{author}{Y.~Guo}, \bibinfo{author}{J.~Zhou},
  \bibinfo{author}{Y.~Gao},
\newblock \bibinfo{title}{Large-scale cross-modality search via collective
  matrix factorization hashing},
\newblock \bibinfo{journal}{IEEE Transactions on Image Processing}
  \bibinfo{volume}{25} (\bibinfo{year}{2016}) \bibinfo{pages}{5427--5440}.
\bibitem[{Kumar and Udupa(2011)}]{94}
\bibinfo{author}{S.~Kumar}, \bibinfo{author}{R.~Udupa},
\newblock \bibinfo{title}{Learning hash functions for cross-view similarity
  search},
\newblock in: \bibinfo{booktitle}{Twenty-second international joint conference
  on artificial intelligence}, \bibinfo{year}{2011}.
\bibitem[{Zhang and Li(2014)}]{95}
\bibinfo{author}{D.~Zhang}, \bibinfo{author}{W.-J. Li},
\newblock \bibinfo{title}{Large-scale supervised multimodal hashing with
  semantic correlation maximization},
\newblock in: \bibinfo{booktitle}{Proceedings of the AAAI conference on
  artificial intelligence}, volume~\bibinfo{volume}{28}, \bibinfo{year}{2014}.
\bibitem[{Lin et~al.(2015)Lin, Ding, Hu, and Wang}]{96}
\bibinfo{author}{Z.~Lin}, \bibinfo{author}{G.~Ding}, \bibinfo{author}{M.~Hu},
  \bibinfo{author}{J.~Wang},
\newblock \bibinfo{title}{Semantics-preserving hashing for cross-view
  retrieval},
\newblock in: \bibinfo{booktitle}{Proceedings of the IEEE conference on
  computer vision and pattern recognition}, \bibinfo{year}{2015}, pp.
  \bibinfo{pages}{3864--3872}.
\bibitem[{Jiang and Li(2017)}]{97}
\bibinfo{author}{Q.-Y. Jiang}, \bibinfo{author}{W.-J. Li},
\newblock \bibinfo{title}{Deep cross-modal hashing},
\newblock in: \bibinfo{booktitle}{Proceedings of the IEEE conference on
  computer vision and pattern recognition}, \bibinfo{year}{2017}, pp.
  \bibinfo{pages}{3232--3240}.
\bibitem[{Yang et~al.(2017)Yang, Deng, Liu, Liu, Tao, and Gao}]{98}
\bibinfo{author}{E.~Yang}, \bibinfo{author}{C.~Deng}, \bibinfo{author}{W.~Liu},
  \bibinfo{author}{X.~Liu}, \bibinfo{author}{D.~Tao}, \bibinfo{author}{X.~Gao},
\newblock \bibinfo{title}{Pairwise relationship guided deep hashing for
  cross-modal retrieval},
\newblock in: \bibinfo{booktitle}{proceedings of the AAAI Conference on
  Artificial Intelligence}, volume~\bibinfo{volume}{31}, \bibinfo{year}{2017}.
\bibitem[{Xie et~al.(2020)Xie, Deng, Li, Liu, and Tao}]{99}
\bibinfo{author}{D.~Xie}, \bibinfo{author}{C.~Deng}, \bibinfo{author}{C.~Li},
  \bibinfo{author}{X.~Liu}, \bibinfo{author}{D.~Tao},
\newblock \bibinfo{title}{Multi-task consistency-preserving adversarial hashing
  for cross-modal retrieval},
\newblock \bibinfo{journal}{IEEE Transactions on Image Processing}
  \bibinfo{volume}{29} (\bibinfo{year}{2020}) \bibinfo{pages}{3626--3637}.
\bibitem[{Lu et~al.(2017)Lu, Chen, and Li}]{61}
\bibinfo{author}{X.~Lu}, \bibinfo{author}{Y.~Chen}, \bibinfo{author}{X.~Li},
\newblock \bibinfo{title}{Hierarchical recurrent neural hashing for image
  retrieval with hierarchical convolutional features},
\newblock \bibinfo{journal}{IEEE transactions on image processing}
  \bibinfo{volume}{27} (\bibinfo{year}{2017}) \bibinfo{pages}{106--120}.
\bibitem[{Yang et~al.(2017)Yang, Lin, and Chen}]{35}
\bibinfo{author}{H.-F. Yang}, \bibinfo{author}{K.~Lin}, \bibinfo{author}{C.-S.
  Chen},
\newblock \bibinfo{title}{Supervised learning of semantics-preserving hash via
  deep convolutional neural networks},
\newblock \bibinfo{journal}{IEEE transactions on pattern analysis and machine
  intelligence} \bibinfo{volume}{40} (\bibinfo{year}{2017})
  \bibinfo{pages}{437--451}.
\bibitem[{Yao et~al.(2016)Yao, Long, Mei, and Rui}]{36}
\bibinfo{author}{T.~Yao}, \bibinfo{author}{F.~Long}, \bibinfo{author}{T.~Mei},
  \bibinfo{author}{Y.~Rui},
\newblock \bibinfo{title}{Deep semantic-preserving and ranking-based hashing
  for image retrieval.},
\newblock in: \bibinfo{booktitle}{IJCAI}, volume~\bibinfo{volume}{1},
  \bibinfo{year}{2016}, p.~\bibinfo{pages}{4}.
\bibitem[{Chen and Lu(2020)}]{38}
\bibinfo{author}{Y.~Chen}, \bibinfo{author}{X.~Lu},
\newblock \bibinfo{title}{Deep discrete hashing with pairwise correlation
  learning},
\newblock \bibinfo{journal}{Neurocomputing} \bibinfo{volume}{385}
  (\bibinfo{year}{2020}) \bibinfo{pages}{111--121}.
\bibitem[{Wang et~al.(2016)Wang, Shi, and Kitani}]{39}
\bibinfo{author}{X.~Wang}, \bibinfo{author}{Y.~Shi}, \bibinfo{author}{K.~M.
  Kitani},
\newblock \bibinfo{title}{Deep supervised hashing with triplet labels},
\newblock in: \bibinfo{booktitle}{Asian conference on computer vision},
  \bibinfo{organization}{Springer}, \bibinfo{year}{2016}, pp.
  \bibinfo{pages}{70--84}.
\bibitem[{Krizhevsky et~al.(2009)Krizhevsky, Hinton et~al.}]{40}
\bibinfo{author}{A.~Krizhevsky}, \bibinfo{author}{G.~Hinton}, et~al.,
\newblock \bibinfo{title}{Learning multiple layers of features from tiny
  images}  (\bibinfo{year}{2009}).
\bibitem[{Lin et~al.(2014)Lin, Maire, Belongie, Hays, Perona, Ramanan,
  Doll{\'a}r, and Zitnick}]{102}
\bibinfo{author}{T.-Y. Lin}, \bibinfo{author}{M.~Maire},
  \bibinfo{author}{S.~Belongie}, \bibinfo{author}{J.~Hays},
  \bibinfo{author}{P.~Perona}, \bibinfo{author}{D.~Ramanan},
  \bibinfo{author}{P.~Doll{\'a}r}, \bibinfo{author}{C.~L. Zitnick},
\newblock \bibinfo{title}{Microsoft coco: Common objects in context},
\newblock in: \bibinfo{booktitle}{European conference on computer vision},
  \bibinfo{organization}{Springer}, \bibinfo{year}{2014}, pp.
  \bibinfo{pages}{740--755}.
\bibitem[{Zhang et~al.(2015)Zhang, Lin, Zhang, Zuo, and Zhang}]{41}
\bibinfo{author}{R.~Zhang}, \bibinfo{author}{L.~Lin},
  \bibinfo{author}{R.~Zhang}, \bibinfo{author}{W.~Zuo},
  \bibinfo{author}{L.~Zhang},
\newblock \bibinfo{title}{Bit-scalable deep hashing with regularized similarity
  learning for image retrieval and person re-identification},
\newblock \bibinfo{journal}{IEEE Transactions on Image Processing}
  \bibinfo{volume}{24} (\bibinfo{year}{2015}) \bibinfo{pages}{4766--4779}.
\bibitem[{Zhang et~al.(2014)Zhang, Zhang, Li, and Guo}]{43}
\bibinfo{author}{P.~Zhang}, \bibinfo{author}{W.~Zhang}, \bibinfo{author}{W.-J.
  Li}, \bibinfo{author}{M.~Guo},
\newblock \bibinfo{title}{Supervised hashing with latent factor models},
\newblock in: \bibinfo{booktitle}{Proceedings of the 37th international ACM
  SIGIR conference on Research \& development in information retrieval},
  \bibinfo{year}{2014}, pp. \bibinfo{pages}{173--182}.
\bibitem[{Zheng et~al.(2020)Zheng, Zhang, and Lu}]{44}
\bibinfo{author}{X.~Zheng}, \bibinfo{author}{Y.~Zhang},
  \bibinfo{author}{X.~Lu},
\newblock \bibinfo{title}{Deep balanced discrete hashing for image retrieval},
\newblock \bibinfo{journal}{Neurocomputing}  (\bibinfo{year}{2020}).
\bibitem[{Yuan et~al.(2020)Yuan, Wang, Zhang, Tay, Jie, Liu, and Feng}]{63}
\bibinfo{author}{L.~Yuan}, \bibinfo{author}{T.~Wang},
  \bibinfo{author}{X.~Zhang}, \bibinfo{author}{F.~E. Tay},
  \bibinfo{author}{Z.~Jie}, \bibinfo{author}{W.~Liu},
  \bibinfo{author}{J.~Feng},
\newblock \bibinfo{title}{Central similarity quantization for efficient image
  and video retrieval},
\newblock in: \bibinfo{booktitle}{Proceedings of the IEEE/CVF Conference on
  Computer Vision and Pattern Recognition}, \bibinfo{year}{2020}, pp.
  \bibinfo{pages}{3083--3092}.
\bibitem[{Kingma and Ba(2015)}]{45}
\bibinfo{author}{D.~P. Kingma}, \bibinfo{author}{J.~Ba},
\newblock \bibinfo{title}{Adam: A method for stochastic optimization},
\newblock \bibinfo{journal}{ICLR (Poster)}  (\bibinfo{year}{2015}).
\bibitem[{Donahue et~al.(2013)Donahue, Jia, Vinyals, Hoffman, and Darrell}]{46}
\bibinfo{author}{J.~Donahue}, \bibinfo{author}{Y.~Jia},
  \bibinfo{author}{O.~Vinyals}, \bibinfo{author}{J.~Hoffman},
  \bibinfo{author}{T.~Darrell},
\newblock \bibinfo{title}{Decaf: A deep convolutional activation feature for
  generic visual recognition}  (\bibinfo{year}{2013}).
\bibitem[{Selvaraju et~al.(2020)Selvaraju, Cogswell, Das, Vedantam, Parikh, and
  Batra}]{64}
\bibinfo{author}{R.~R. Selvaraju}, \bibinfo{author}{M.~Cogswell},
  \bibinfo{author}{A.~Das}, \bibinfo{author}{R.~Vedantam},
  \bibinfo{author}{D.~Parikh}, \bibinfo{author}{D.~Batra},
\newblock \bibinfo{title}{Grad-cam: Visual explanations from deep networks via
  gradient-based localization},
\newblock \bibinfo{journal}{International Journal of Computer Vision}
  \bibinfo{volume}{128} (\bibinfo{year}{2020}) \bibinfo{pages}{336--359}.

\end{thebibliography}



\bio{figs/a1}
\textbf{Zhengyang Yu} received his B.S. degree in the College of Computer \& Information Science, Southwest University. His current research interests include Multimedia Search, Large-scale Instance Retrieval and Person Re-Identification.  
\endbio

\vspace{40pt}

\bio{figs/a4}
\textbf{Song Wu} received his B.S. degree and M.S. degree in Computer Science from the Southwest University,
Chongqing, China, in 2009 and 2012, respectively. He received his Ph.D from the Leiden Institute of Advanced
Computer Science (LIACS), Leiden University, Netherlands.
He is a member of the Overseas High-level Talent Program in Chongqing and currently working at the College of Computer and Information Science of Southwest University. His current research interests include large-scale image retrieval and classification, big data technology and deep learning based computer vision (the co-author of the most cited paper of journal Neurocomputing: Deep learning for visual understanding: A review).
\endbio

\vspace{2pt}

\bio{figs/a2}
\textbf{Zhihao Dou} received his B.S degree in Automation in the College of Computer \& Information Science, Southwest University in 2020. He works as an intern in State Grid Corporation of China. His research interests are computer version and data mining.
\endbio

\vspace{40pt}

\bio{figs/a3}
\textbf{Erwin M. Bakker} is co-director of the LIACS Media Lab at Leiden University. He has published widely in the fields
of image retrieval, audio analysis and retrieval and bioinformatics. He was closely involved with the start of the International Conference on Image and Video Retrieval (CIVR) serving on the organizing committee in 2003 and
2005. Moreover, he regularly serves as a program committee member or organizing committee member for scientific multimedia and human-computer interaction conferences and workshops.
\endbio

\end{document}